\documentclass[letterpaper]{article} 
\usepackage{aaai25}  
\usepackage{times}  
\usepackage{helvet}  
\usepackage{courier}  
\usepackage[hyphens]{url}  
\usepackage{graphicx} 
\usepackage{natbib}  
\usepackage{caption} 
\usepackage{microtype}
\usepackage{epsfig}
\usepackage{algpseudocode}
\usepackage{algorithm2e}
\usepackage{amsmath,amssymb}
\usepackage[listofformat=parens, justification=centering, subrefformat=subparens]{subfig}

\newcommand\etal[1]{\textit{et al.}~\cite{#1}}

\newcommand\Caption[3][]{\caption[#2]{\label{#1}\textsc{\small#2}. \small#3}}

\newcommand\fig[1]{Fig.~\ref{fig:#1}}

\newcommand\tab[1]{Tab.~\ref{tab:#1}}

\DeclareMathOperator*{\argmax}{arg\,max}

\algnewcommand\algorithmicforeach{\textbf{for each}}
\algdef{S}[FOR]{ForEach}[1]{\algorithmicforeach\ #1\ \algorithmicdo}

\usepackage{tikz}
\usetikzlibrary{calc}
\usepackage{rotating}
\usepackage{multirow,tabularx}
\newcolumntype{?}{!{\vrule width 1.5pt}}

\newif\ifarxiv
\arxivtrue

\ifarxiv
\usepackage{fancyhdr}
\fi

\title{GHOST: Gaussian Hypothesis Open-Set Technique}

\author {
    Ryan Rabinowitz\textsuperscript{\rm 1},
    Steve Cruz\textsuperscript{\rm 2},
    Manuel Günther\textsuperscript{\rm 3},
    and Terrance E. Boult\textsuperscript{\rm 1}
}
\affiliations {
    \textsuperscript{\rm 1}Vision and Security Technology Lab, University of Colorado Colorado Springs,\\
    \textsuperscript{\rm 2}Computer Vision Research Lab, University of Notre Dame,\\
    \textsuperscript{\rm 3}Department of Informatics, University of Zurich\\
    \{rrabinow, tboult\}@uccs.edu,
    stevecruz@nd.edu,
    guenther@ifi.uzh.ch

}

\ifarxiv
{
  \chead{\footnotesize This is a pre-print of the original paper accepted at the AAAI Conference on Artificial Intelligence 2025.}
  \lhead{}
  \thispagestyle{fancy}
}
\fi

\begin{document}
\ifarxiv
\nocopyright
\fi

\maketitle

\begin{abstract}
Evaluations of large-scale recognition methods typically focus on overall performance.
While this approach is common, it often fails to provide insights into performance across individual classes, which can lead to fairness issues and misrepresentation.
Addressing these gaps is crucial for accurately assessing how well methods handle novel or unseen classes and ensuring a fair evaluation.
To address fairness in Open-Set Recognition (OSR), we demonstrate that per-class performance can vary dramatically.
We introduce Gaussian Hypothesis Open Set Technique (GHOST), a novel hyperparameter-free algorithm that models deep features using class-wise multivariate Gaussian distributions with diagonal covariance matrices.
We apply Z-score normalization to logits to mitigate the impact of feature magnitudes that deviate from the model’s expectations, thereby reducing the likelihood of the network assigning a high score to an unknown sample.
We evaluate GHOST across multiple ImageNet-1K pre-trained deep networks and test it with four different unknown datasets.
Using standard metrics such as AUOSCR, AUROC and FPR95, we achieve statistically significant improvements, advancing the state-of-the-art in large-scale OSR.
Source code is provided online.

\end{abstract}
\begin{links}
    \link{Code Repository}{https://github.com/Vastlab/GHOST}
\end{links}

%
\section{Introduction}
\label{sec:intro}

When deploying deep neural networks (DNNs) in real-world environments, they must handle a wide range of inputs.
The ``closed-set assumption," prevalent in most evaluations, represents a significant limitation of traditional recognition-oriented machine learning algorithms \cite{scheirer2012toward}.
This assumption presumes that the set of possible classes an algorithm will encounter is known a priori, meaning that these algorithms are not evaluated for robustness against samples from previously unseen classes.
Open-Set Recognition (OSR) challenges this assumption by requiring designs that anticipate encountering samples from unknown classes during testing.

\begin{figure}[t!]
  \centerline{\includegraphics[width=0.95\columnwidth]{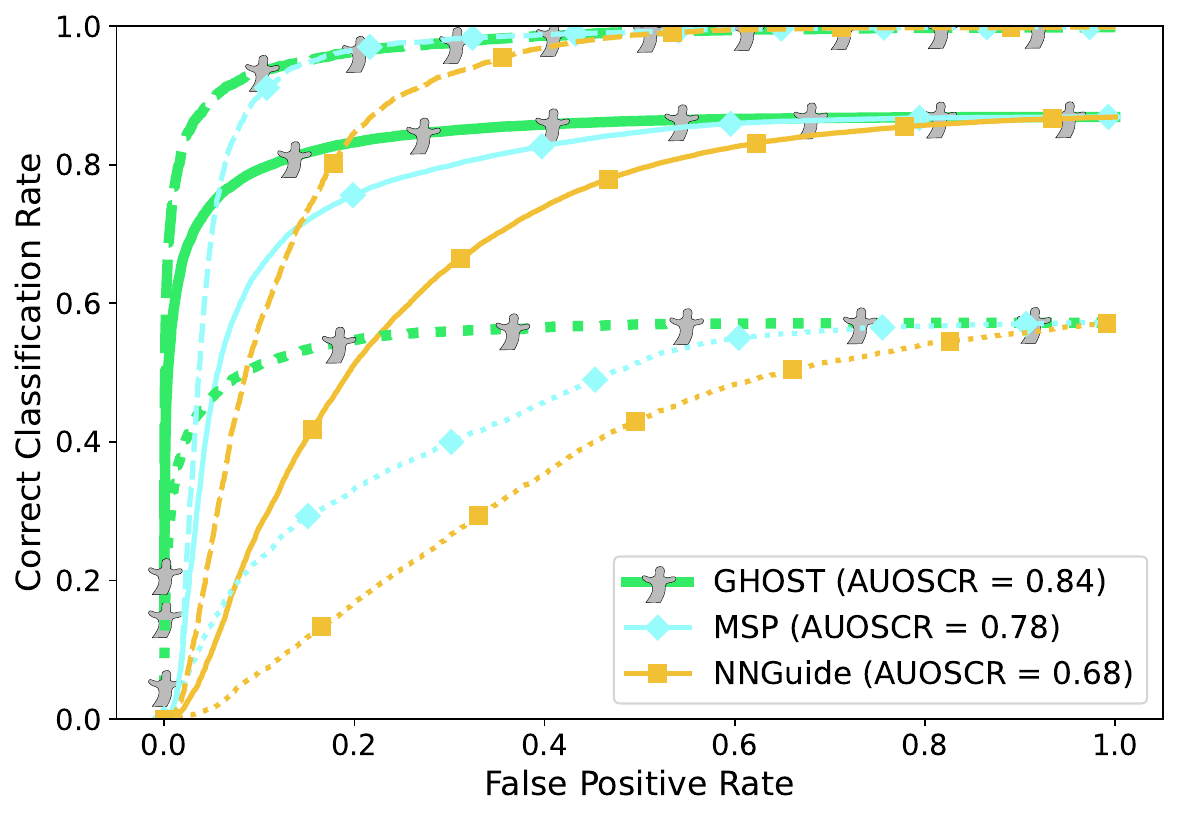}}
  \Caption[fig:teaser]{Class-wise Open-Set Recognition}{OSCR comparison using the MAE-H architecture with OpenImage-O as unknowns. Overall performance is the solid line; Average performance on easy (top 10\%)  and hard (bottom 10\%) classes shown as dashed/dotted lines, respectively. We compare GHOST with Maximum Softmax Probability (MSP) and  NNGuide.  Also, we show the area under the curve (AUC) of each method's overall OSCR. GHOST outperforms in each setting and maintains its correct classification rate as the FPR rate decreases while others fall off dramatically; hence, GHOST maintains fairness in difficult cases while improving overall OSR. 
  }
\end{figure}

Often, OSR is performed by thresholding on confidence \cite{hendrycks17baseline,vaze2022openset} or having an explicit ``other'' class \cite{ge2017gopenmax} and computing overall performance, ignoring the effects of per-class performance differentials \cite{li2023accurate}.
However, evaluating recognition systems under OSR conditions is crucial for understanding their behavior in real-world scenarios.
This paper shows that as more unknowns are rejected, there is great variation in per-class accuracy, which could lead to unfair treatment of underperforming classes, see \fig{teaser}.

Recently, research has followed two primary methodologies for adapting DNNs to OSR problems: (1) training processes that enhance feature spaces and (2) post-processing techniques applied to pre-trained DNNs to adjust their outputs for identifying known and unknown samples \cite{roady2020open}.
Although OSR training methods have occasionally proven effective \cite{zhou_learning_2021, miller2021class, dhamija2018reducing}, their application is complex due to the evolving nature of DNNs and the specific, often costly training requirements for each.
If different DNNs are trained in various ways, why should a single OSR training technique be universally applicable?
Furthermore, if an OSR technique is specific to a particular DNN, its value diminishes as state-of-the-art DNNs evolve.
In contrast, post-processing methods, such as leveraging network embeddings \cite{bendale2016openmax}, are more straightforward to implement and can be applied to almost any DNN.
These methods avoid the complexities associated with training techniques and focus instead on evaluating performance.
Thus, the challenge becomes: \textbf{\emph{how can various DNNs designed with a closed-set assumption be adapted for OSR problems?}}

Initial post-processing OSR algorithms \cite{bendale2016openmax, rudd2017evm} used distance metrics in high-dimensional feature spaces to relate inference samples to known class data from training.
However, choosing appropriate hyperparameters, such as a distance metric, is not straightforward, particularly for networks trained without distance metric learning, leading to an expensive parameter search.
Further, large-scale datasets like ImageNet \cite{deng2009imagenet} and small-scale splits \cite{neal2018counterfactual, perera_generative-discriminative_2020, yang2020convolutional, geng2020recent, zhou_learning_2021} often lack suitable train-validation-test splits for a fair parameter search.

Additionally, a major limitation with prior evaluations is their emphasis on overall performance, rather than ensuring robust performance for each individual class.
This focus can obscure significant disparities between classes, leading to an incomplete understanding of the algorithm's effectiveness and potentially resulting in unfair treatment of some classes.
For example, an algorithm might achieve high overall accuracy but fail to recognize rare or challenging classes, which is critical for applications requiring high precision across all classes.
Such evaluations can misrepresent the algorithm’s performance for underrepresented or underperforming classes, which may be overlooked when only aggregate metrics are considered.
This lack of detailed analysis can lead to skewed evaluations, where the model’s weaknesses in specific areas are not addressed, ultimately affecting its real-world applicability and fairness.
While fairness is not a major concern for ImageNet-1K, the dataset used herein, we consider it a reasonable proxy for operational open-set problems due to its size and widespread use as a feature extractor or for fine-tuning domain-specific models.

We propose a novel post-processing OSR algorithm, the Gaussian Hypothesis Open-Set Technique (GHOST), which uses per-class multivariate Gaussian models with diagonal covariance of DNN embeddings to reduce network overconfidence for unknown samples.
The use of per-class modeling is crucial for ensuring fairness across all classes.
By modeling each feature dimension separately for each class, GHOST evaluates each class on its own merits, rather than grouping them together.
This technique helps address the challenge of handling the worst-performing classes fairly and reduces the risk of the model being overly confident about samples from these difficult classes.
Importantly, GHOST eliminates the need for hyperparameters, simplifying the application of OSR techniques for end-users.
Our novel GHOST algorithm improves traditional OSR measures and fairness, achieving a win-win outcome in line with recent fairness goals presented by \citet{islam2021can,li2023accurate}.

\noindent\textbf{In summary, our main contributions are}:
\begin{itemize}
    \item We introduce GHOST, a novel, state-of-the-art, hyper\-parameter-free post-processing algorithm that models per-feature, per-class distributions to improve per-class OSR.
    \item We present an extensive experimental analysis that adapts both the previous and recent state-of-the-art methods while evaluating multiple state-of-the-art DNNs, with results showing that GHOST is statistically significantly better on both global OSR and OOD metrics.
    \item We provide the first fairness analysis in OSR, identify significant per-class differences in large-scale OSR evaluations, and demonstrate that GHOST improves fairness.
\end{itemize}

\section{Related Work}
\label{sec:related}

Some methods have been proposed to improve the training of DNNS for OSR \cite{zhang2022learning,xu2023contrastive,wan2024unlocking,wang2024exploring,li2024all, li2024prototype, sensoy2018evidential}.
We do not consider these as direct competitors, as they go beyond statistical inference and train reconstruction models and use generative techniques or other additional training processes.  
Post-processing methods, including GHOST, can all use better features, but as \citet{vaze2022openset} pointed out, better closed-set classifiers improve performance more and are continuing to evolve rapidly, so our focus is on post-processing algorithms.   

Post-hoc approaches are well-explored in out-of-distribution (OOD) detection. 
Moreover, they are used in various practical settings requiring large pre-trained networks. 
The first attempt to adapt pre-trained DNNs for OSR using statistical inference on representations extracted from a pre-trained backbone was made by \citet{bendale2016openmax}.
They sought to replace the popular SoftMax layer, which is problematic for OSR, with OpenMax.
OpenMax computes the centroid for each known class from training data and uses Extreme Value Theory to fit Weibull distributions over the distance from the centroid to the training samples.
During inference, the probabilities that a sample belongs to a known class are converted to probabilities of unknown, which are summed and effectively form an additional category representing the probability of unknown.
The Extreme Value Machine (EVM) proposed by \citet{rudd2017evm} is another OSR system based on statistical inference using distance between samples. 
It finds a set of extreme vectors in each training-set class and fits a Weibull distribution on the distance between them and the closest samples of other ``negative" classes in high-dimensional feature space.
Both systems compute distances in high-dimensional space, so a practitioner must select a distance metric that applies to their DNN backbone. 
This process often requires a search over possible metrics and other algorithm-related hyperparameters.
We might consider these methods to be direct competitors as they employ straightforward statistical measures to recognize known samples, but large scale evaluation shows they are not as effective as some simple baselines \cite{bisgin2024large}. 

Using network outputs to reject unknowns is widely used, and \citet{hendrycks17baseline,hendrycks2022scaling} showed that thresholding on Maximum  Softmax Probability (MSP) or Maximum Logits (MaxLogit) from a closed-set DNN provides good baselines for OSR. 
In addition, \citet{vaze2022openset} went so far as to argue that good closed-set classifiers with logit-based thresholding are sufficient for OSR. 
We also consider the popular energy-based OOD detection technique \cite{liu2020energy}, which computes energy based on the logit vector of a DNN (this method's performance is subpar, and so it is relegated to the supplemental). 
A recent collection of OOD methods, OpenOOD \cite{yang2022openood,zhang2023openood}, has compared many of these post-hoc methods using recent, large-scale datasets.
Herein, we consider only the best performing: Nearest Neighbor Guidance (NNGuide) \cite{park2023nearest} for OOD detection (others in the supplemental).
This method scales the confidence output from a DNN's classification layer by the sample's cosine similarity to a subset of training data, and is currently leading in the OpenOOD ImageNet-1K leaderboard\footnote{\url{http://zjysteven.github.io/OpenOOD}} and so we use it as a primary comparison.
We show that GHOST normalization, which does not need a reference set, improves performance overall, setting a new standard for large-scale OSR and OOD.

\section{Approach}
\subsection{A Gaussian Hypothesis}
\label{sec:approach}

The first works on open-set recognition and open-set deep networks \cite{scheirer2012toward,scheirer2014probability,bendale2016openmax,rudd2017evm} all focused on the most distant points within a class or the values at the class boundaries.  
Hence, it is natural that they employed extreme-value theory as their underlying model. 
Having evaluated many of these EVT-based approaches in practical settings, we found a few significant difficulties: These methods are sensitive to outliers/mislabeled data (due to their reliance on a small percentage of extreme data) and have a high cost and sensitivity of tuning their hyperparameters.  
A final difficulty with this approach is reducing the high-dimensional features into a 1-dimensional distance, typically Euclidean or Cosine.

Features within a DNN are learned using large amounts of data.   
Various papers have shown that, with some mild assumptions, convergence in a two-layer network follows a central-limit theory \cite{sirignano2020mean}, and using a mean-field analysis that was extended to some older deep network architectures \cite{lu2020mean} -- so there are inherently some reasons to hypothesize Gaussian models. 

We start by summarizing the main simple NN central-limit theory of \cite{sirignano2020mean}, which indicates that for a large number $M$ of neurons, the empirical distribution of the neural network's parameters behaves as  Gaussian distribution. 
Their theorems show that, given their assumptions, the empirical distribution of the parameters behaves as a Gaussian distribution with a specific variance-covariance structure.  
Central to the proofs of these theories is mean-field theory, and the convergence of the parameters to the mean follows from the central limit theorem. These mean-field distributional convergence results were then extended to some older deep networks \cite{lu2020mean}, but extending to new networks is complex.  
We believe that empirical testing, as we do in our experiments, is a sufficient and much easier way to evaluate the Gaussian hypothesis for any new network.

Inspired by those theories, we hypothesize that similarly, when input is from a class seen in training, each value in the network can be reasonably approximated by a multivariate Gaussian and that, importantly, out-of-distribution samples would be more likely to be inconsistent with the resulting Gaussian model.  
While the theories of \citet{sirignano2020mean,lu2020mean} are about the learnable network parameters, we hypothesize that with a Gaussian distribution per parameter, after many layers of computation, for a set of inputs from a given class, the distribution of each embedding value may also be well modeled with a Gaussian. 
Critical in this hypothesis is that for at least the embedding $\vec\varphi$ as shown in \fig{processing}, which is used to compute the per-class logits, Gaussian models are class-specific;  \fig{GHOST} shows an example model with sample values for a known and outlier. 

Due to the complexities and variations of modern DNN architectures, formally proving that this hypothesis is valid for every DNN is impractical and unlikely.  Instead, we derive a technique from this hypothesis and apply it to the most well-performing, publicly available architectures to prove its utility.

\begin{figure}[t]
  \small\centering
  \begin{tikzpicture}[scale=.5]
    \node[draw](Image) at (0,0) {\includegraphics[width=.1\textwidth]{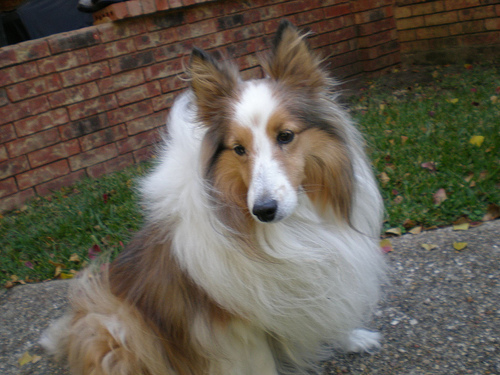}};
    \node[draw,rectangle,minimum height=2cm](Backbone) at (4,0) {
        \begin{turn}{270}\begin{minipage}{1.5cm}\Large \centering Backbone\\ Network\end{minipage}\end{turn}
    };
    \node[draw,rectangle](Features) at (6.5,0) {\begin{turn}{270} Embeddings $\vec \varphi$\end{turn}};
    \node[draw,rectangle](Logits) at (10.3,0) {\begin{turn}{270} Logits $\vec z$\end{turn}};
    \node[draw,rectangle](Probabilities) at (13,0) {\begin{turn}{270} Confidences $\vec y$\end{turn}};


    \draw[->,thick] (Image) -> (Backbone);
    \draw[->,thick] (Backbone) -> (Features);
    \draw[->,thick] (Features) -> node [above] {\begin{turn}{270} Linear \end{turn}} node [below]{\begin{turn}{270} $\mathbf W$\end{turn}} (Logits);
    \draw[->,thick] (Logits) -> node [above] {\begin{turn}{270} SoftMax\end{turn}} (Probabilities);

    \node[draw, rectangle, minimum height=1.5cm, minimum width=0.9cm] (GHOST) at (8,-4.5) {\begin{turn}{270}\begin{minipage}{1.5cm}\centering Gaussians\\[1ex]$\{(\vec\mu_k,\vec\sigma_k)\}$ \end{minipage}\end{turn}};
     \node[draw, rectangle, minimum height=1.5cm, minimum width=0.5cm] (Scoring) at (10.3,-4.5) {\begin{turn}{270} GHOST\end{turn}};

    \node[draw, rectangle, minimum height=1.5cm, minimum width=.9cm] (OOD) at (12.5,-4.5) {\begin{turn}{270}\begin{minipage}{1cm} \centering GHOST score $\gamma$ \end{minipage}\end{turn}};

    \draw[->,thick,dashed] (Features) |- (GHOST);
    \draw[->,thick,dashed] (Logits) -- (Scoring);
    \draw[->,thick,dashed] (GHOST) -- (Scoring);
    \draw[->,thick,dashed] (Scoring) -- (OOD);

  \end{tikzpicture}
  \Caption[fig:processing]{GHOST Scores}{In a pre-trained network indicated with solid arrows, an image is presented to the backbone network, which extracts deep feature embeddings $\vec\varphi$ that are then processed with a Linear layer to logits $\vec z$, and further with SoftMax to probabilities $\vec y$. 
    For training GHOST, we extract embeddings from training data, from which we model class-wise multivariate Gaussian distributions. 
    During evaluation, the Gaussian of the predicted class are used to turn the embeddings $\vec\varphi$ into z-score, which is used together with the maximum logit $z_{\hat k}$ to compute the GHOST score $\gamma$.}
\end{figure}

\begin{figure*}[bth]
    \centerline{\includegraphics[width=0.85\textwidth]{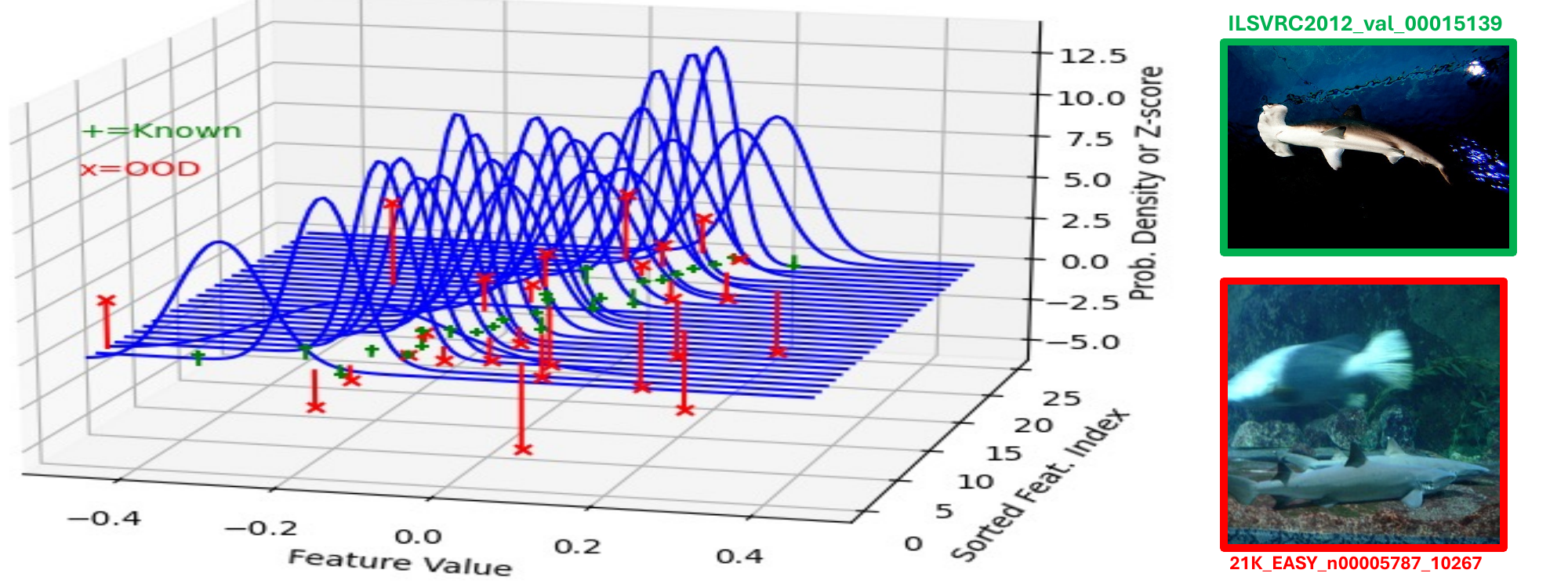}}
  \Caption[fig:GHOST]{GHOST modeling of a Multivariate Gaussian per Class}{Samples of Gaussians from the MAE-H network are shown on the left, sampled once every 30 dimensions.  Dimensions were sorted on mean value to improve visibility, and the spread shows how some dimensions have greater variance than others.    The plot also shows the value of per-dimension z-scores associated with a correctly classified hammerhead image (known in green) and an OOD example with a shark (red) misclassified as a hammerhead.  The z-scores of the OOD example are much larger than those of the known.  More examples in supplemental material. }
\end{figure*}

\paragraph{GHOST Training.} Consider the network processing shown in \fig{processing}.
Given a training dataset $\mathcal X=\{(x_n,t_n)\mid 1\leq n\leq N\}$ with $N$ samples $x_n$ and their class labels $1\leq t_n\leq K$ representing $K$ known classes.
Here, we apply post-processing, so we assume the backbone network to be trained on the same $K$ classes contained in $\mathcal X$.
For each correctly-classified training sample, we use the backbone to extract its $D$-dimensional embedding $\vec\varphi\in\mathbb R^D$.
For each class $k$, we model a multivariate Gaussian distribution with mean $\vec \mu_k$ and diagonal covariance $\vec\sigma_k$ from the samples of that class and collect these Gaussians for all classes as $\mathcal G = \{(\vec\mu_k, \vec\sigma_k)\mid 1\leq k\leq K\}$ via:
\begin{equation}
    \begin{aligned}
    \vec\mu_k = \frac1{N_k}&\sum\limits_{(x_n,t_n)\in\mathcal X} \mathbb I(k,t_n)\cdot \vec\varphi_n  \\ 
    \vec\sigma_k^2 = \frac1{N_k-1}&\sum\limits_{(x_n,t_n)\in\mathcal X} \mathbb I(k,t_n)\cdot(\vec\mu_k-\vec\varphi_n)^2
    \end{aligned}
\end{equation}
where the indicator function $\mathbb I(k,t_n)$ is 1 if the sample $x_n$ belongs to class $k$ and is correctly classified,\footnote{This restriction reduces the influence of mislabeled samples.} and $N_k$ is the number of correctly-classified samples for class $k$.
Hence, each feature dimension of each known class is modeled from its training data.
Together, these Gaussian models are useful for differentiating unknown samples.

\paragraph{GHOST Inference.} Building from the open-set theory by \citet{bendale2016openmax}, we know a provably open-set algorithm is produced if the confidence scores are monotonically decreasing with the distance of a feature from a mean.
To achieve this with our Gaussian Hypothesis, we combine our model with the intuition that there are significant deviations of DNN's embedding magnitude when an unknown sample is encountered \cite{dhamija2018reducing,cruz2024oosa} and, thus, the embedding $\vec\varphi$ deviates from all class means, even though the angular direction might overlap with a certain class' mean $\vec\mu_k$.
For a given test sample, we first compute embedding $\vec\varphi$, logits $\vec z$ and the predicted class $\hat k$ as:
\begin{equation}
    \label{eq:max-class}
    \hat k = \argmax_k z_k\,.
\end{equation}
We select the associated Gaussian $(\vec\mu_{\hat k}, \vec\sigma_{\hat k})$ to compute our z-score:
\begin{equation}
    s = \sum_{d=1}^D \frac{|\varphi_d - \mu_{\hat k, d}|}{\sigma_{\hat k, d}}
\end{equation}
which is small if the embedding is close to -- and larger the more it deviates from -- the mean.
Unlike Euclidean or cosine distance to reduce dimensionality in a geometrically fixed way, this z-score-based deviation measure adapts to the inherent ``shape'' of the embeddings differently for each class.
Some classes may have large variations in some dimension $\varphi_d$, whereas others have minor variations.  

The most obvious way to use the z-score $s$ to ensure monotonically decreasing score and generate an open-set algorithm is by dividing the predicted class' logit $z_{\hat k}$:
\begin{equation}
  \gamma = \frac{z_{\hat k}}{s}\,.
\end{equation}
If the sample is close to the class mean of the predicted class $\hat k$, $\gamma$ increases in scale, whereas a large z-score $s$ lead to a reduction of $\gamma$.
Thus, thresholding on $\gamma$ to reject items as unknown or out-of-distribution is consistent with formal open-set theory.

Note that we are not normalizing the predicted $\gamma$ score, but basically threshold this score directly, comparable to MaxLogit \cite{hendrycks2022scaling}.
As compared to OpenMax, an advantage of GHOST is that the Gaussian models $\mathcal G$ are less sensitive to outliers or mislabeled data in the training set, as the contribution of any input is only to the mean and standard deviation, which reduces noise.  
In contrast, even a single outlier can dominate the computation of Weibulls  \cite{scheirer2012toward,scheirer2014probability,bendale2016openmax,rudd2017evm}. 
Furthermore, GHOST removes the necessity of selecting a tail size for Weibull fitting, and there is no need to choose a distance metric.

\subsection{Class-Based Evaluation}
\label{sec:metrics}
When evaluating the performance of Open-Set Recognition, we make use of the Open-set Classification Rate (OSCR) \cite{dhamija2018reducing} as our primary metric since it was specifically designed to evaluate OSR performance at a given operational threshold.
While OSCR was designed for Open-Set Recognition and the effect of unknown samples, it is related to Accuracy-Rejection Curves which examine performance of systems with respect to uncertainty of new samples from known classes\cite{nadeem2009accuracy}.
We split our test dataset in known samples $\mathcal K = \{(x_n, t_n)\mid 1\leq n\leq N_K\}$ and unknown samples $\mathcal U = \{(x_n)\mid 1\leq n \leq N_U\}$ that do not have class labels.
An OSCR plot shows the Correct Classification Rate (CCR) versus the False Positive Rate (FPR):
\begin{equation}
  \begin{aligned}
      \mathrm{CCR}(\theta)&= \frac{\bigl|\{(x_n,t_n) \in \mathcal K \wedge \hat k = t_n \wedge \gamma \geq \theta \}\bigr|}{|\mathcal K|}\,,\\
      \label{eq:oscr}
      \mathrm{FPR}(\theta)&= \frac{\bigl|\{x_n \in \mathcal U \wedge \gamma \geq \theta \}\bigr|}{|\mathcal U|}\,.
  \end{aligned}
\end{equation}
For other algorithms, we replace $\gamma$ by $z_{\hat k}$ (MaxLogit), $y_{\hat k}$ (MSP), or other prediction scores for maximum class $\hat k$.
To plot the CCR at a specific FPR=$\tau$, one can invert the FPR to compute $\theta_{\tau}=\mathrm{FPR}^{-1}(\tau)$, which allows us to line up different algorithms/classes.
We also utilize the area under the OSCR curve (AUOSCR) to compare algorithms across all thresholds, but we wish to emphasize that this suffers from many of the same problems as AUROC because it combines all possible thresholds, which is not how systems operate.   

A fact that is overlooked by all OSR evaluations is the difference in performance for different classes.
Only few researchers evaluated the variance of closed-set accuracy across classes.
Since this is related to algorithmic fairness, we go a step further and compute the variances and coefficients of variation of CCR values across classes.
First, we split our test dataset into samples from certain classes $\mathcal K_k = \{x_n\mid (x_n,k) \in \mathcal K\}$ and compute per-class CCR at FPR $\tau$:
\begin{equation}
  \label{eq:per-class}
  \mathrm{CCR}_k(\theta_\tau) = \frac{\bigl|\{x_n \in \mathcal K_k \wedge \hat k = k \wedge \gamma \geq \theta_\tau \}\bigr|}{|\mathcal K_k|}
\end{equation}
Note that we do not compute per-class thresholds/FPR here;  the same set of thresholds $\theta_\tau$ is used in all classes, but is different for each algorithm.
We follow the idea of \citet{atkinson1970measurement,formby1999coefficient,xinying2023guide} and compute the mean,  variance and coefficient of variation of per-class CCR at FPR=$\tau$:
\begin{equation}
    \begin{aligned}
    \label{eq:variation}
    \mu_{\mathrm{CCR}}(\theta_\tau) = \frac1K &\sum\limits_k \mathrm{CCR}_k (\theta_\tau)\\ 
    \sigma^2_{\mathrm{CCR}}(\theta_\tau) = \frac1{K-1} &\sum\limits_k \bigl(\mathrm{CCR}_k(\theta_\tau) - \mu_{\mathrm{CCR}}(\theta_\tau)\bigr)^2 \\
    {\cal V}_{\mathrm{CCR}}(\theta_\tau) &= \frac{\sigma_{\mathrm{CCR}}(\theta_\tau)} {\mu_{\mathrm{CCR}}(\theta_\tau)}
    \end{aligned}
\end{equation}
where ${\cal V}_{\mathrm{CCR}}$ provides a commonly used measure of inequality (unfairness) that facilitates comparisons by normalizing for changes in mean values.
We evaluate $\mu_{\mathrm{CCR}}, \sigma_{\mathrm{CCR}}$ and ${\cal V}_{\mathrm{CCR}}$ at various FPR values. 
Since the CCR at FPR=1 represents closed-set accuracy, ${\cal V}_{\mathrm{CCR}}(\theta_1)$ corresponds to the unfairness in closed-set accuracy, while the associated $\mu_{\mathrm{CCR}}(\theta_1)$ represents closed-set accuracy (because the same number of samples exist per known class). 
To highlight some of the performance differentials, we also sort the classes by their closed-set accuracy and show the average CCR over the top-10 and bottom-10 percent of classes.

In prior OSR works, the Area Under the Receiver Operating Characteristic (AUROC) has been identified as an important metric for OSR evaluations.
Binary unknown rejection is rather conceptually related to OOD, but we include AUROC as a secondary metric and additionally present ROC curves in the supplemental material.
For both metrics, the curves themselves must accompany reported Area Under-statistics because area alone cannot distinguish if competing curves cross and characterize performance at specific ranges of sensitivities. 
We present more OSCR curves in the supplement.
Additionally, in a problem where 90\% of data (or risk) comes from potentially unknown inputs, having very low FPR is important and may not be easily discernable in linear plots.
For highlighting differences for high-security applications, we follow \citet{dhamija2018reducing} and plot OSRC and ROC curves with logarithmic x-axes (as in \fig{log}).
For additional quantitative insight into high-security performance, we report FPR95 in our overall results (\tab{overall_metrics}).

We also introduce a new OSR measure that avoids integrating over thresholds.
Our goal is to determine the lowest FPR that maintains a set classification accuracy level.
We call this F@C95, where we compute the FPR at the point where CCR is 95\% of the closed-set accuracy.
This measure, analogous to FPR95 in binary out-of-distribution detection, uses CCR and can be applied overall or per class.
\section{Experiments}
\label{sec:experiments}
Our main evaluation relies on large-scale datasets that cover both known and unknown samples.
Specifically, we draw from recent large-scale settings \cite{vaze2022openset,hendrycks17baseline,bitterwolf2023or} that differ from other evaluations which use only small-scale data with few classes and low-resolution images.
We include additional results in the supplemental material.

In particular, we use ImageNet-1K \cite{ILSVRC15} pre-trained networks and the validation set as our test set for knowns.
For unknowns, we consider multiple datasets from the literature.
We utilize a recent purpose-built OOD dataset called No ImageNet Class Objects (NINCO) \cite{bitterwolf2023or}, which consists of images specifically excluding any semantically overlapping or background ImageNet objects.
Additionally, we use the ImageNet-21K-P Open-Set splits (\emph{Easy} \& \emph{Hard}) proposed by \citet{vaze2022openset} in their semantic shift benchmark.
We also include OpenImage-O \cite{wang2022vim}, a dataset constructed from a public image database. 
Further details on each dataset and comparisons on additional datasets such as Places \cite{zhou2017places}, SUN \cite{xiao2010sun}, and Textures \cite{Cimpoi_2014_CVPR} are provided in the supplemental.

\subsection{Experimental Setup}
\paragraph{Methods.} We compare GHOST with Maximum Logit (MaxLogit) \cite{hendrycks2022scaling,vaze2022openset}, which is currently the state-of-the-art in large-scale OSR according to \citet{vaze2022openset}, and Maximum Softmax Probability (MSP) \cite{hendrycks17baseline,vaze2022openset}.
For completeness, we also compare with the current state-of-the-art in large-scale OOD according to the OpenOOD \cite{yang2022openood,zhang2023openood} leaderboard,\footnote{\url{https://zjysteven.github.io/OpenOOD}} NNGuide \cite{park2023nearest}.
Note that NNGuide has been adapted to more recent architectures than those used in their paper, as we have observed that this adaptation significantly impacts performance.
Also, GHOST results on OpenOOD's ImageNet-1K benchmark are found in the supplemental, as well as a comparison with SCALE \cite{xu2024scaling}, REACT \cite{sun2021react} and KNN \cite{sun2022out}.

\paragraph{Architectures.} We utilize two architectures: Masked Auto Encoder-trained Vision Transformer MAE-H \cite{he2022masked} and ConvNeXtV2-H \cite{woo2023convnext}.
MAE-H is a ViT-H network trained with a masked autoencoder; it is competitive with the state-of-the-art, PeCo \cite{dong2023peco}, which does not have publicly available code or checkpoints.
ConvNeXtV2-H is a recent, high-performing convolutional neural network (CNN). 
It is included to show that GHOST performance gains are not limited to transformer-based networks.
Both networks were trained exclusively with ImageNet-1K by their respective authors.
We report results on additional networks in the supplemental, offering evidence for generalizability to other architectures.

\section{Results and Discussion}
\begin{table*}[]
\centering
\small
\resizebox{\linewidth}{!}{
\begin{tabular}{l?cccccccc}
\multicolumn{1}{c?}{\multirow{3}{*}{\textbf{\begin{tabular}[c]{@{}c@{}}Unknowns\\ \end{tabular}}}} & \multicolumn{8}{c}{\textbf{$\uparrow$ AUOSCR}  \textbf{$\uparrow$ AUROC} \textbf{$\downarrow$ FPR95}} \\ \cline{2-9} 
\multicolumn{1}{c?}{} & \multicolumn{4}{c?}{\textbf{MAE-H}} & \multicolumn{4}{c}{\textbf{ConvNeXtV2-H}} \\ \cline{2-9} 
\multicolumn{1}{c?}{} & \multicolumn{1}{c|}{\begin{tabular}[c]{@{}c@{}}\textbf{GHOST} (ours)\end{tabular}} & \multicolumn{1}{c|}{\begin{tabular}[c]{@{}c@{}}\textbf{MSP} \end{tabular}} & \multicolumn{1}{c|}{\begin{tabular}[c]{@{}c@{}}\textbf{MaxLogit} \end{tabular}} & \multicolumn{1}{c?}{\begin{tabular}[c]{@{}c@{}}\textbf{NNGuide} \end{tabular}} & \multicolumn{1}{c|}{\begin{tabular}[c]{@{}c@{}}\textbf{GHOST} (ours)\end{tabular}} & \multicolumn{1}{c|}{\begin{tabular}[c]{@{}c@{}}\textbf{MSP} \end{tabular}} & \multicolumn{1}{c|}{\begin{tabular}[c]{@{}c@{}}\textbf{MaxLogit} \end{tabular}} & \multicolumn{1}{c}{\begin{tabular}[c]{@{}c@{}}\textbf{NNGuide} \end{tabular}} \\ \noalign{\hrule height 1.5pt}
21K-P \emph{Easy} & \multicolumn{1}{c|}{$\uparrow$ \textbf{.75} $\uparrow$ \textbf{.84} $\downarrow$ \textbf{.58}} & \multicolumn{1}{c|}{$\uparrow$ .72 $\uparrow$ .80 $\downarrow$ .65} & \multicolumn{1}{c|}{$\uparrow$ .67 $\uparrow$  .75 $\downarrow$ .63} & \multicolumn{1}{c?}{$\uparrow$ .62 $\uparrow$ .69 $\downarrow$ .80} & \multicolumn{1}{c|}{$\uparrow$ \textbf{.74} $\uparrow$ \textbf{.83} $\downarrow$ \textbf{.60}} & \multicolumn{1}{c|}{$\uparrow$ .72 $\uparrow$ .79 $\downarrow$ .65} & \multicolumn{1}{c|}{$\uparrow$ .68 $\uparrow$  .75 $\downarrow$ .64} & $\uparrow$ .70 $\uparrow$ .79 $\downarrow$ .70 \\
21K-P \emph{Hard} & \multicolumn{1}{c|}{$\uparrow$ \textbf{.73} $\uparrow$ \textbf{.81} $\downarrow$ \textbf{.62}} & \multicolumn{1}{c|}{$\uparrow$ .69 $\uparrow$ .75 $\downarrow$ .75} & \multicolumn{1}{c|}{$\uparrow$ .65 $\uparrow$  .71 $\downarrow$ .74} & \multicolumn{1}{c?}{$\uparrow$ .47 $\uparrow$ .52 $\downarrow$ .89} & \multicolumn{1}{c|}{$\uparrow$ \textbf{.72} $\uparrow$ \textbf{.80} $\downarrow$ \textbf{.65}} & \multicolumn{1}{c|}{$\uparrow$ .68 $\uparrow$ .74 $\downarrow$ .76} & \multicolumn{1}{c|}{$\uparrow$ .65 $\uparrow$  .72 $\downarrow$ .74} & $\uparrow$ .60 $\uparrow$ .67 $\downarrow$ .83 \\
NINCO & \multicolumn{1}{c|}{$\uparrow$ \textbf{.81} $\uparrow$ \textbf{.91} $\downarrow$ \textbf{.47}} & \multicolumn{1}{c|}{$\uparrow$ .78 $\uparrow$ .83 $\downarrow$ .65} & \multicolumn{1}{c|}{$\uparrow$ .73 $\uparrow$  .79 $\downarrow$ .62} & \multicolumn{1}{c?}{$\uparrow$ .49 $\uparrow$ .55 $\downarrow$ .88} &             \multicolumn{1}{c|}{$\uparrow$ \textbf{.79} $\uparrow$ \textbf{.89} $\downarrow$ \textbf{.50}} & \multicolumn{1}{c|}{$\uparrow$ .75 $\uparrow$ .83 $\downarrow$ .64} & \multicolumn{1}{c|}{$\uparrow$ .73 $\uparrow$  .82 $\downarrow$ .60} & $\uparrow$ .74 $\uparrow$ .74 $\downarrow$ .78 \\
OpenImage-O & \multicolumn{1}{c|}{$\uparrow$ \textbf{.84} $\uparrow$ \textbf{.95} $\downarrow$ \textbf{.26}} & \multicolumn{1}{c|}{$\uparrow$ .76 $\uparrow$ .87 $\downarrow$ .52} & \multicolumn{1}{c|}{$\uparrow$ .71 $\uparrow$  .82 $\downarrow$ .49} & \multicolumn{1}{c?}{$\uparrow$ .68 $\uparrow$ .77 $\downarrow$ .64} &       \multicolumn{1}{c|}{$\uparrow$ \textbf{.83} $\uparrow$ \textbf{.94} $\downarrow$ \textbf{.32}} & \multicolumn{1}{c|}{$\uparrow$ .79 $\uparrow$ .88 $\downarrow$ .49} & \multicolumn{1}{c|}{$\uparrow$ .77 $\uparrow$  .87 $\downarrow$ .44} & $\uparrow$ .66 $\uparrow$ .83 $\downarrow$ .64
\end{tabular}}
\Caption[tab:overall_metrics]{Overall Quantitative Results}{On two state-of-the-art pre-trained architectures, GHOST achieves new state-of-the-art performance across all three metrics. In the supplemental, we demonstrate that the improvements provided by GHOST are statistically significant and consistent across additional unknowns and architectures. Methods such as Energy, SCALE, and others that are less effective are found in the supplemental.}
\end{table*}

\begin{table}
\centering
\small
\resizebox{\linewidth}{!}{
\begin{tabular}{@{}l?c|c|c|c|c@{}}
\textbf{Unknowns} & \textbf{GHOST} & \textbf{MSP} & \textbf{MaxLogit} & \textbf{NNGuide} & \textbf{Energy} \\ \noalign{\hrule height 1.5pt}
21K-P \emph{Easy}   & \textbf{0.48}  & 0.53         & 0.54              & 0.76             & 0.65            \\
21K-P \emph{Hard}   & \textbf{0.53}  & 0.64         & 0.64              & 0.86             & 0.74            \\
NINCO        & \textbf{0.35}  & 0.51         & 0.51              & 0.86             & 0.62            \\
OpenImage-O & \textbf{0.17}  & 0.39         & 0.39              & 0.60             & 0.50           
\end{tabular}}
\Caption[tab:mae-fatc]{F@C95}{The corresponding minimum FPR at 95\% of closed set accuracy ($\downarrow$). Each method's results are computed on a pre-trained MAE-H. Energy is shown here as space permitted, but additional comparisons with Energy are in the supplemental.}
\end{table}

\begin{figure}[tb]
  \centerline{\includegraphics[width=0.95\columnwidth]{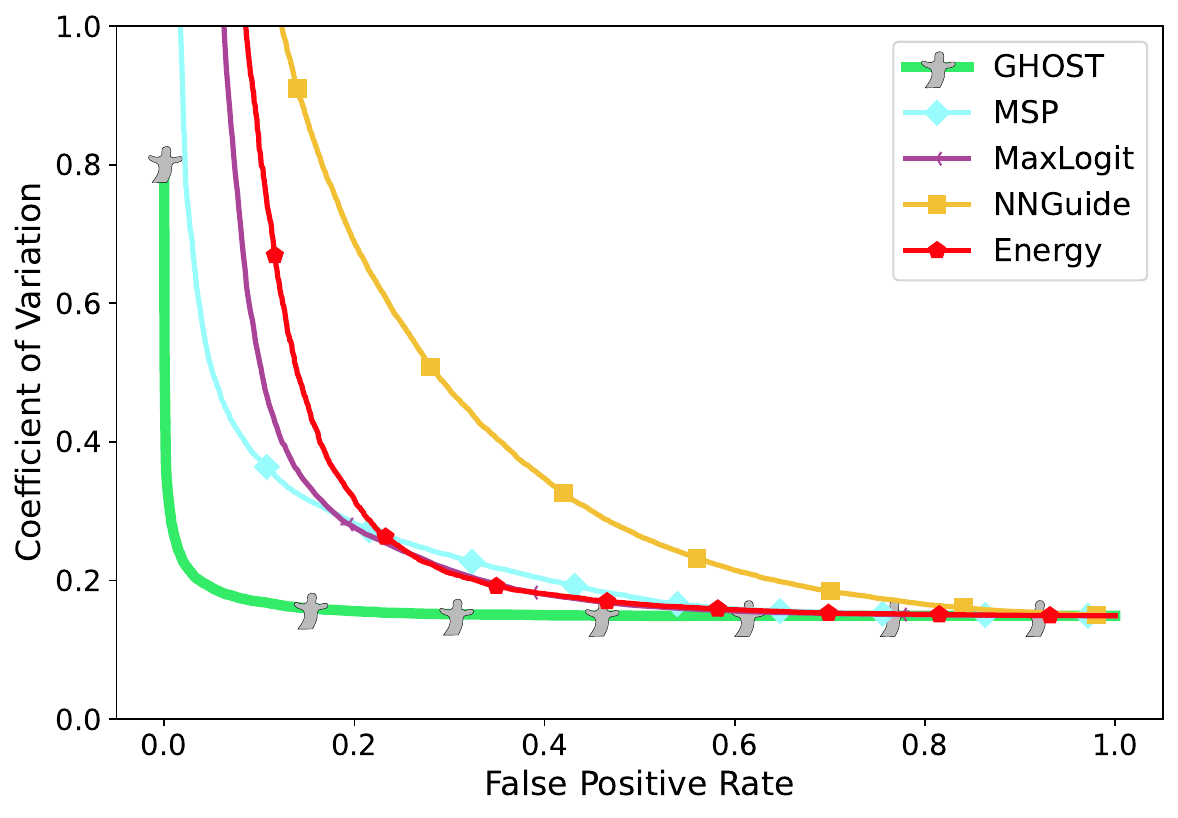}}
  \Caption[fig:unfair]{Unfairness (Coefficient of Variation)}{This figure shows the unfairness of OSR algorithms across False Positive Rates using MAE-H network with OpenImages as unknowns. All algorithms include the inherent unfairness from the base classifier on the far right, but GHOST maintains its level much better as FPR rates are decreased to the left.
    }
\end{figure}

\paragraph{Global Performance.}
We present some of our quantitative results in \tab{overall_metrics}, while more datasets are found in the supplemental material.
On open-set AUOSCR, \tab{overall_metrics} shows that GHOST outperforms other methods on all datasets with an absolute gain of at least 2\%.
While OOD is not the primary focus, GHOST also outperforms other methods on the AUROC measure (\tab{overall_metrics}) with a lead of 4\%. 
It is important to note that on the NINCO \cite{bitterwolf2023or} dataset, which was specifically designed to avoid overlap with ImageNet-1K, GHOST shows clear and convincing performance gains in terms of both AUROC and AUOSCR, and some of the reduced performance for others may be a sign of overlap. 
We present results on our proposed F@C95 in \tab{mae-fatc}, where each method reports the minimum effective FPR it achieves while maintaining 95\% of the closed-set accuracy. 
On each dataset, GHOST achieves far lower F@C95 rates than other methods.

Naturally, statistical testing should be used to validate the hypothesis of superior performance.  
To this end, we present statistical analysis in the supplemental material and summarize it here.
For AUROC, GHOST very significantly outperforms all methods on all datasets with $p<10^{-6}$.
For AUOSCR, GHOST is significantly better overall and on most datasets with $p<10^{-3}$.

\begin{table}
\centering
\small
\resizebox{\linewidth}{!}{
\begin{tabular}{@{}l?c|c|c|c|c@{}}
\textbf{Unknowns} & \textbf{GHOST} & \textbf{MSP} & \textbf{MaxLogit} & \textbf{NNGuide} & \textbf{Energy} \\ \noalign{\hrule height 1.5pt}
21K-P \emph{Easy} & \textbf{0.32}  & 0.55         & 0.68              & 1.35             & 0.83            \\
21K-P \emph{Hard} & \textbf{0.36}  & 0.60         & 0.61              & 2.28             & 0.69          \\
NINCO & \textbf{0.21}  & 0.45         & 0.50              & 2.18             & 0.68            \\
OpenImage-O & \textbf{0.17}  & 0.38         & 0.52              & 1.16             & 0.82    
\end{tabular}}
\Caption[tab:mae-cov]{Coefficient of Variance}{The unfairness measure ${\cal V}_{\mathrm{CCR}}$ coefficients ($\downarrow$) of all methods. Each is computed on a pre-trained MAE-H at 10\% FPR and evaluated on various unknown datasets. Energy is shown here as space permitted, but additional comparisons with Energy are in the supplement.}
\end{table}

\begin{figure}[t] 
\centerline{\includegraphics[width=0.95\columnwidth]
{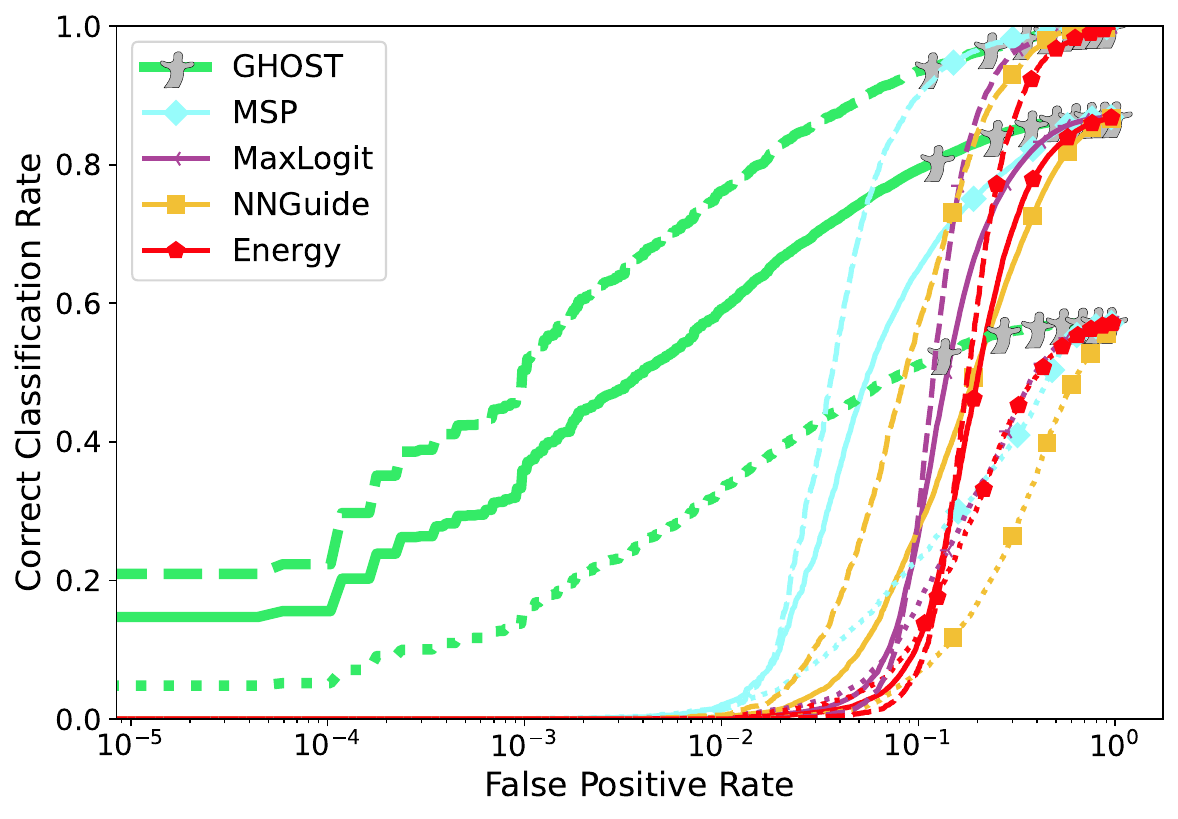}}
  \Caption[fig:log]{OSCR in Logscale}{In applications with the high cost of false-positives or those with many potential unknowns, it is more important to focus on low FPR performance, in which case log FPR as shown here are more useful. The global performance is presented as a solid line, while top-10\,\% is dashed, and bottom-10\,\% is dotted.  In all cases, GHOST is significantly better at low FPR levels, and below FPR of 0.1 GHOST's bottom-10\,\% performance is better than most algorithms' top-10\,\%.}
\end{figure}

\paragraph{Fairness and Class-Based Evaluation.}
Fairness in OSR has two components -- the inherent differential ability of the base network and the ability of the OSR to maintain the accuracy of classes in a fair/balanced manner.  
Previous work has ignored how individual classes are impacted by OSR thresholding.  
In \fig{unfair}, we use the ${\cal V}_{\mathrm{CCR}}$ coefficient of \eqref{eq:variation}.
As this is a common measure of unfairness, lower values are more equitable.  
At the right-hand side for FPR=1, we see the network's baseline unfairness, and while GHOST maintains that for the majority of lower FPR values, other algorithms quickly degrade.  
Notably, the traditional ``recognition'' algorithms MSP/MaxLogit maintain their fairness better than the OOD-oriented algorithms, though none are even close to GHOST at lower FPRs.   
Given that we cannot compute the area under this curve as many of the curves are not bounded, we show quantitative values at 10\% FPR in \tab{mae-cov}.

To provide a more detailed analysis on the class differentials, we plot the OSCR for different classes.
For each algorithm, we select the top-10\,\% of best-performing and bottom-10\,\% of worst-performing classes based on the closed-set class-based accuracy, which is identical to CCR$_k(\theta_1)$ as defined in \eqref{eq:per-class}.
We combine these classes and plot OSCR curves.
In \fig{teaser} and \fig{log}, these best- and worst-performing classes are shown together with the global OSCR curve that includes all classes.
While \fig{teaser} presents a linear FPR-axis, \fig{log} shows a logarithmic FPR axis that allows investigation of very low FPR ranges.
Especially in \fig{teaser} it is obvious that with decreasing FPR, GHOST provides the same drop of CCR for all three lines, whereas other algorithms have different behavior.
Especially MSP has superior CCR for well-classified classes (top-10\,\%), while dropping much quicker for difficult bottom-10\,\% classes.
Furthermore, \fig{log} shows that this behavior of GHOST extends to very low FPRs, levels that are not even reached by other methods.

From our results in \tab{overall_metrics} and \tab{mae-cov}, GHOST dominates performance across the board, in both well accepted OSR and OOD metrics (AUOSCR, FPR95, AUROC) as well as in class-wise fairness.
We present additional OSCR and ROC curves in the supplemental for interested readers to verify the consistency of our results on area-based metrics.

\subsection{Testing the Gaussian Hypothesis}
To empirically test the Gaussian hypothesis, we use the Shapiro-Wilk test for normality with Holm's step-down procedure for family-wise error control \cite{trawinski2012nonparametric}. 
For MAE-H and ConvNeXtV2-H, pretrained networks, only 2.79\% and 3.13\% of per-class distributions rejected the null hypothesis (normality), consistent with expected 95\% confidence. 
Tests on Swin-T and DenseNet-121 rejected normality for 0.56\% and 8.27\% of the distributions, indicating the Gaussian assumption does not hold for every class in every network, but it still generally holds. 
We include the results for these additional networks in the supplemental material.

Additionally, GHOST could be adapted to use full covariance matrices and mahalanobis distance, but we leave this adaptation for future work.
\section{Conclusion}
\label{sec:conclusion}
In this paper, we propose GHOST, our Gaussian Hypothesis Open-Set Technique, which follows the formal definition of provable open-set theory for deep networks, based on the theorems by \citet{bendale2016openmax}. 
We hypothesize that using per-class, per-dimension Gaussian models of feature vectors to normalize raw network logits can effectively differentiate unknown samples and improve OSR performance.    
Although this remains a hypothesis, it may be valuable for future work to explore mean-field theory as a means to formally prove it. 
By utilizing Gaussian models, we move away from traditional assumptions that rely on distance metrics in high-dimensional spaces.
Instead, we normalize logits through a sum of z-scores.
These Gaussian models are more robust to outliers, which can significantly affect extreme value-based statistics.
We demonstrate this on two distinct architectures, providing strong support for our assumption.

Our experiments provide compelling evidence, setting a new state-of-the-art performance. 
Using both networks, we achieve superior results in AUOSCR and AUROC with ImageNet-1K as knowns and four datasets (and more in the supplemental) as unknowns. 
In nearly all cases, GHOST outperforms all methods, with performance gains being statistically very significant (shown in the supplemental).
Furthermore, GHOST is computationally efficient and easy to store, requiring only the mean and standard deviation (i.e., two floats per feature per class). 
A pre-trained network requires just one pass over the validation or training data to compute the GHOST model, and its test-time complexity is $\mathcal{O}(1)$. 

We are the first to investigate fairness in OSR by examining class-wise performance differences, and we hope to encourage research that incorporates more fairness-related metrics for OSR.
We have shown that GHOST maintains the closed-set unfairness of the original classifier across most FPRs, whereas other algorithms struggle significantly, increasing unfairness even at moderate FPRs.

\clearpage
\twocolumn[\begin{minipage}{.99\textwidth}\centering%
    \LARGE\bf GHOST: Gaussian Hypothesis Open-Set Technique --- Supplemental Material\vspace*{2ex}

    \Large Ryan Rabinowitz, Steve Cruz, Manuel G\"unther, and Terrance E. Boult\vspace*{3ex}
\end{minipage}]

\begin{figure*}[t]
  \centering 
  \includegraphics[width=0.8\textwidth]{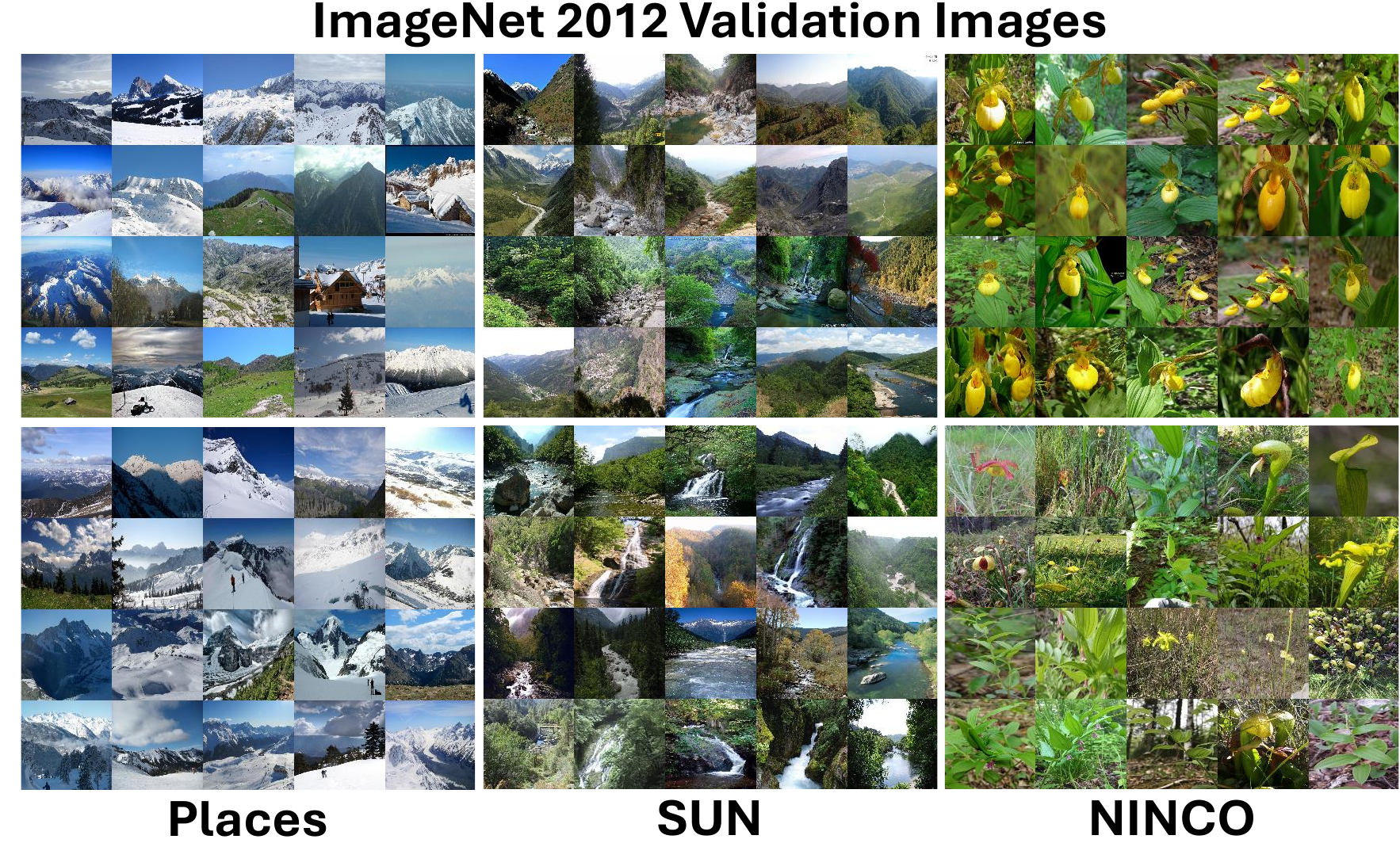}
  \Caption[fig:NINCO_Examples]{Misclassified Unknown Samples}{We examined the unknown samples most frequently mispredicted from three OOD datasets: Places \cite{zhou2017places} SUN \cite{xiao2010sun}, and NINCO \cite{bitterwolf2023or}. We merged the 20 most confident samples into collages and created counterparts from the ImageNet 2012 validation set. Both Places and SUN are known to have significant class overlap \cite{bitterwolf2023or}, but NINCO was specifically designed to avoid such issues. We can see NINCO's most confused class merely has a similar background to a known ImageNet class, whereas Places and SUN seemingly have direct overlap. This separation between known and unknown data for ImageNet-1K/NINCO reinforces the significance of our results on NINCO.}
\end{figure*}

\section{Dataset details}

\subsubsection{ImageNet}
ImageNet-2012 \cite{ILSVRC15} consists of 1000 classes, each with up to 1600 images in training and 50 images in validation.
Images were collected from internet search queries and verified through Amazon Mechanical Turk. 
ImageNet classes largely consist of animals, manufactured artifacts, and some natural structures, foods, and plants. 
Many Images are taken from a person-level perspective and, due to the nature of internet-based harvesting, collected from various sources, locations, and cameras. 
The diverse large-scale nature of this dataset has made it central to image recognition; it has become a common benchmark for new networks. 
We utilize ImageNet-2012 as our knowns dataset because, in addition to its scale and complexity, many pre-trained models are available, allowing us to compare performance across various networks.

\subsubsection{Datasets from OOD Literature}
We utilize a recent purpose-built OOD dataset, NINCO: No ImageNet Class Objects \cite{bitterwolf2023or}, that was built from images that specifically exclude any semantically overlapping or background ImageNet objects.
While NINCO is not as large as ImageNet-2012 val, its authors took extensive measures to reduce data contamination.
We also utilize OpenImage-O \cite{wang2022vim}, a dataset constructed from a public image database.
We present results on other, older OOD datasets in which NINCO alleged contamination as part of an extended evaluation.

\subsubsection{Datasets from OSR Literature}
\label{sec:datasets}
ImageNet 21K-P \emph{Easy}/\emph{Hard} Open-Set Splits were proposed by Vaze \etal{vaze2022openset} in their semantic shift benchmark. 
Each split contains 50K images from ImageNet-21K \cite{ridnik2021}, and base classes were selected using semantic WordNet distance from ImageNet-1K.
We do not utilize the other semantic shift splits proposed by \cite{vaze2022openset}, as they are small-scale and would require retraining networks.

\section{Statistical Analysis and Additional Datasets}

We perform statistical testing by sampling each dataset (and the val set) 10 times, selecting 1000 knowns and unknowns.   
We use paired two-tailed t-tests to evaluate the statistical significance of each method's performance compared to GHOST. 
We find statistical significance across the board, except on SUN and Places, when GHOST was compared to MSC see \tab{full_auroc_sig}.

\begin{table*}[p]
\Caption[tab:full_auroc_sig]{Statistical Significance Evaluation}{AUROC paired t-tests results on OOD and OSR datasets specified in the main paper and additional OOD datasets SUN, iNaturalist, Textures, Places using MAE-H or ConvNeXtV2-H as the pretrained network. P-values are from two-sided paired t-test, with Bonferroni corrections,  for rejecting the null hypothesis that the given algorithm performance is the same as GHOST. }\centering
\subfloat[MAE-H]{
\begin{tabular}{lll|l|l|l|l}
             &                                    & GHOST & MSC                    & MaxLogit               & NNGuide         & Energy           \\ \cline{2-7} 
NINCO        & \multicolumn{1}{l|}{Mean}          & 0.900 & 0.825                  & 0.782                  & 0.543           & 0.738            \\
             & \multicolumn{1}{l|}{STD}           & 0.004 & 0.008                  & 0.009                  & 0.008           & 0.009            \\
             & \multicolumn{1}{l|}{P-Value}       &       & 6.838e-12              & 3.241e-12              & 2.48e-15        & 9.790e-13        \\ \hline
OpenImage\_O & \multicolumn{1}{l|}{Mean}          & 0.956 & 0.874                  & 0.822                  & 0.772           & 0.771            \\
             & \multicolumn{1}{l|}{STD}           & 0.005 & 0.010                  & 0.010                  & 0.011           & 0.013            \\
             & \multicolumn{1}{l|}{P-Value}       &       & 4.564e-10              & 2.178e-12              & 4.4605e-13      & 2.133e-12        \\ \hline
Places       & \multicolumn{1}{l|}{Mean}          & 0.830 & 0.823                  & 0.756                  & 0.803           & 0.694            \\
             & \multicolumn{1}{l|}{STD}           & 0.009 & 0.009                  & 0.011                  & 0.009           & 0.012            \\
             & \multicolumn{1}{l|}{P-Value}       &       & 0.003                  & 1.212e-10              & 0.0001          & 5.142e-12        \\ \hline
SUN          & \multicolumn{1}{l|}{Mean}          & 0.851 & 0.835                  & 0.784                  & 0.804           & 0.733            \\
             & \multicolumn{1}{l|}{STD}           & 0.009 & 0.009                  & 0.012                  & 0.012           & 0.013            \\
             & \multicolumn{1}{l|}{P-Value}       &       & 0.0002                 & 7.805e-10              & 9.106e-07       & 1.038e-11        \\ \hline
Textures     & \multicolumn{1}{l|}{Mean}          & 0.916 & 0.868                  & 0.857                  & 0.505           & 0.842            \\
             & \multicolumn{1}{l|}{STD}           & 0.004 & 0.008                  & 0.008                  & 0.004           & 0.008            \\
             & \multicolumn{1}{l|}{P-Value}       &       & 3.491e-09              & 7.601e-10              & 6.722e-19       & 8.784e-11        \\ \hline
easy\_21k    & \multicolumn{1}{l|}{Mean}          & 0.837 & 0.796                  & 0.747                  & 0.694           & 0.701            \\
             & \multicolumn{1}{l|}{STD}           & 0.011 & 0.009                  & 0.012                  & 0.011           & 0.011            \\
             & \multicolumn{1}{l|}{P-Value}       &       & 1.480e-08              & 1.798e-10              & 5.0781e-11      & 8.432e-12        \\ \hline
hard\_21k    & \multicolumn{1}{l|}{Mean}          & 0.812 & 0.748                  & 0.717                  & 0.522           & 0.683            \\
             & \multicolumn{1}{l|}{STD}           & 0.012 & 0.010                  & 0.009                  & 0.013           & 0.007            \\
             & \multicolumn{1}{l|}{P-Value}       &       & 7.909e-10              & 6.672e-11              & 9.607e-14       & 2.880e-11        \\ \hline
iNaturalist  & \multicolumn{1}{l|}{Mean}          & 0.971 & 0.910                  & 0.889                  & 0.759           & 0.848            \\
             & \multicolumn{1}{l|}{STD}           & 0.004 & 0.007                  & 0.007                  & 0.008           & 0.006            \\
             & \multicolumn{1}{l|}{P-Value}       &       & 3.412e-12              & 3.286e-13              & 9.297e-15       & 1.664e-14 
\end{tabular}
}

\subfloat[ConvNeXtV2-H]{
\begin{tabular}{lll|l|l|l|l}
             &                                      & GHOST                  & MSC                    & MaxLogit               & NNGuide        & Energy         \\ \cline{2-7} 
NINCO        & \multicolumn{1}{l|}{Mean}            & 0.887                  & 0.828                  & 0.816                  & 0.737          & 0.800          \\
             & \multicolumn{1}{l|}{STD}             & 0.004                  & 0.005                  & 0.005                  & 0.009          & 0.006          \\
             & \multicolumn{1}{l|}{P-Value}         &                        & 4.057e-11              & 2.233e-10              & 6.248e-13      & 1.580e-10      \\ \hline
OpenImage\_O & \multicolumn{1}{l|}{Mean}            & 0.940                  & 0.884                  & 0.869                  & 0.830          & 0.849          \\
             & \multicolumn{1}{l|}{STD}             & 0.004                  & 0.006                  & 0.007                  & 0.007          & 0.007          \\
             & \multicolumn{1}{l|}{P-Value}         &                        & 1.552e-11              & 4.927e-12              & 6.118e-12      & 1.051e-11      \\ \hline
Places       & \multicolumn{1}{l|}{Mean}            & 0.827                  & 0.832                  & 0.789                  & 0.830          & 0.757          \\
             & \multicolumn{1}{l|}{STD}             & 0.008                  & 0.007                  & 0.008                  & 0.009          & 0.009          \\
             & \multicolumn{1}{l|}{P-Value}         &                        & 0.003                  & 1.219e-08              & 0.349          & 9.834e-11      \\ \hline
SUN          & \multicolumn{1}{l|}{Mean}            & 0.847                  & 0.846                  & 0.810                  & 0.845          & 0.779          \\
             & \multicolumn{1}{l|}{STD}             & 0.014                  & 0.012                  & 0.012                  & 0.005          & 0.012          \\
             & \multicolumn{1}{l|}{P-Value}         &                        & 0.494                  & 9.629e-08              & 0.646          & 1.002e-08      \\ \hline
Textures     & \multicolumn{1}{l|}{Mean}            & 0.892                  & 0.869                  & 0.876                  & 0.757          & 0.875          \\
             & \multicolumn{1}{l|}{STD}             & 0.006                  & 0.007                  & 0.008                  & 0.011          & 0.008          \\
             & \multicolumn{1}{l|}{P-Value}         &                        & 5.443e-08              & 1.908e-06              & 5.265e-10      & 4.378e-07      \\ \hline
easy\_21k    & \multicolumn{1}{l|}{Mean}            & 0.823                  & 0.791                  & 0.749                  & 0.789          & 0.718          \\
             & \multicolumn{1}{l|}{STD}             & 0.010                  & 0.009                  & 0.011                  & 0.007          & 0.013          \\
             & \multicolumn{1}{l|}{P-Value}         & 1.012e-06              & 7.497e-11              & 8.210e-07              &                & 5.304e-12      \\ \hline
hard\_21k    & \multicolumn{1}{l|}{Mean}            & 0.797                  & 0.736                  & 0.707                  & 0.674          & 0.685          \\
             & \multicolumn{1}{l|}{STD}             & 0.006                  & 0.009                  & 0.012                  & 0.013          & 0.013          \\
             & \multicolumn{1}{l|}{P-Value}         &                        & 6.759e-10              & 2.797e-10              & 3.891e-09      & 9.324e-11      \\ \hline
iNaturalist  & \multicolumn{1}{l|}{Mean}            & 0.965                  & 0.911                  & 0.901                  & 0.882          & 0.881          \\
             & \multicolumn{1}{l|}{STD}             & 0.005                  & 0.005                  & 0.003                  & 0.007          & 0.003          \\
             & \multicolumn{1}{l|}{P-Value}         &                        & 1.270e-10              & 1.480e-11              & 3.513e-11      & 2.035e-11
\end{tabular}
}
\end{table*}

\section{Normality Testing and Z-Score Plots}
\label{shapiro}
We conducted further statistical tests on raw features per-class per-dimension for normality. 
This involved 1,280,000 tests (1280 feature dimensions x 1000 classes) for MAE-H. 
We used the Shapiro-Wilk test for normality and Holm's step-down procedure for Family-wise error control. 
For MAE-H and ConvNeXtV2-H, only 2.79\% and 3.13\% of distributions rejected the null hypothesis (normality), as expected with 95\% confidence. 
Despite different architectures, both networks use autoencoder training. 
To better test effectiveness beyond these networks, we examined two additional influential networks using torchvision ImageNet-1K weights. 
Tests on Swin-T and DenseNet-121 rejected normality 0.56\% and 8.27\% of the distributions indicating the Gaussian assumption does not hold for every network. 
We believe the Shapiro-Wilks test (with Holm's step-down procedure) is a straight-forward method for determining if GHOST should be applied to a given DNN for OSR and urge end-users to investigate their target architecture before applying GHOST.
To this end, we include the code for running this test (and procedure) with our codebase for the convenience of GHOST's end-users.
We provide some insights into the performance of GHOST on DenseNet in a later section (Results on Additional Networks).

\section{OpenOOD}
\label{sec:openood}
We run GHOST on the OpenOOD large-scale ImageNet-1K benchmark.
We consistently find performance on Near-OOD better than Far-OOD tasks across two different networks.
ViT-B-16 achieves an AUROC of 80.83 and 90.75 for Near/Far OOD respectively.
Swin-T achieves an AUROC of 84.97 and 91.09 on Near/Far OOD.
These results show GHOST performs superior to other post-hoc algorithms in Near-OOD using the same backbones.

\section{Results on Additional Networks and Methods}
\label{additional_networks}
We present results on additional networks in \tab{sup_global_metrics} and \tab{coeff-f95c}.
We note the poor performance on DenseNet-121 compared to the other networks we tested.
We find the decrease in performance is consistent with the Shapiro-Wilks statistical testing in \ref{shapiro}, which revealed that 8\% of per-class feature dimensions are not Gaussian for Densenet. 
Further, we recognize that out of MAE-H, ConvNeXtV2-H, Swin-T and DenseNet-121, DenseNet has the lowest closed-set accuracy, with a performance gap of 7\% to the nearest network accuracy. 
Additionally, we note that NNGuide's performance is dramatically worse than on other networks, which may be related to the drop in accuracy.
We leave exploring the relation between Gaussian features and network accuracy to future work.

We want to emphasize, however, even with DenseNet's sub-par accuracy, GHOST maintains a much lower coefficient of  variance than other methods, showing that GHOST's superior fairness still holds, even with a much worse base network.

We also present results from additional OOD post-processors SCALE \cite{xu2024scaling}, REACT \cite{sun2021react}, and KNN \cite{sun2022out} in \tab{sup_global_metrics}. 
For REACT, we found the found the feature activation clipping threshold using the train set rather than using the test set (as the original publication did) for fairness.
While KNN and REACT sometimes perform better than GHOST on the 21K-P Easy dataset, NINCO, the dataset built to avoid overlap with imagenet, still shows GHOST exhibits dramatic performance gains over other methods.
Additionally, paired t-tests reveal GHOST's performance is statistically significantly better across all datasets and both networks when compared with REACT, with P-Values of <0.0135, <0.0066 and <0.047 for AUOSCR, AUROC, and FPR95.
Similarly, GHOST's performance against KNN is statistically significantly better in terms of AUOSCR and AUROC, with P-Values of <0.0285 and <0.0363.
However, GHOST is slightly significantly worse than KNN in terms of FPR95, with a P-Value of <0.047.
\begin{table*}[tp]
\centering
\small
\Caption[tab:sup_global_metrics]{Quantitative Results}{We present quantitative results using AUOSCR, AUROC and FPR95 on four different pre-trained networks. We use the ImageNet-1K-trained checkpoint provided by torchvision or the official checkpoints provided by authors. For MAE-H and ConvNeXtV2-H we present results from additional post-processors SCALE \cite{xu2024scaling}, REACT \cite{sun2021react}, and KNN \cite{sun2022out}.}

\subfloat[Swin-T]{
\resizebox{\linewidth}{!}{
\begin{tabular}{l?ccccc}
\multicolumn{1}{c?}{\multirow{2}{*}{\textbf{\begin{tabular}[c]{@{}c@{}}Unknowns\\ \end{tabular}}}} & \multicolumn{5}{c}{\textbf{$\uparrow$ AUOSCR}  \textbf{$\uparrow$ AUROC} \textbf{$\downarrow$ FPR95}} \\ \cline{2-6} 
\multicolumn{1}{c?}{} & \multicolumn{1}{c|}{\begin{tabular}[c]{@{}c@{}}\textbf{GHOST} (ours)\end{tabular}} & \multicolumn{1}{c|}{\begin{tabular}[c]{@{}c@{}}\textbf{MSP} \end{tabular}} & \multicolumn{1}{c|}{\begin{tabular}[c]{@{}c@{}}\textbf{MaxLogit} \end{tabular}} & \multicolumn{1}{c|}{\begin{tabular}[c]{@{}c@{}}\textbf{NNGuide} \end{tabular}} & \multicolumn{1}{c}{\begin{tabular}[c]{@{}c@{}}\textbf{Energy} \end{tabular}} \\ \noalign{\hrule height 1.5pt}
21K-P \emph{Easy} & \multicolumn{1}{c|}{$\uparrow$ 0.67 $\uparrow$ \textbf{0.78} $\downarrow$ \textbf{0.72}} & \multicolumn{1}{c|}{$\uparrow$ \textbf{0.68} $\uparrow$ \textbf{0.78} $\downarrow$ 0.71} & \multicolumn{1}{c|}{$\uparrow$ 0.64 $\uparrow$  0.74 $\downarrow$ 0.71} & \multicolumn{1}{c|}{$\uparrow$ 0.56 $\uparrow$ 0.66 $\downarrow$ 0.93} & \multicolumn{1}{c}{$\uparrow$ 0.59 $\uparrow$ 0.70 $\downarrow$ 0.77} \\
21K-P \emph{Hard} & \multicolumn{1}{c|}{$\uparrow$ \textbf{0.65} $\uparrow$ \textbf{0.76} $\downarrow$ \textbf{0.77}} & \multicolumn{1}{c|}{$\uparrow$ 0.64 $\uparrow$ 0.72 $\downarrow$ 0.83} & \multicolumn{1}{c|}{$\uparrow$ 0.62 $\uparrow$  0.70 $\downarrow$ 0.82} & \multicolumn{1}{c|}{$\uparrow$ 0.39 $\uparrow$ 0.45 $\downarrow$ 0.97} & \multicolumn{1}{c}{$\uparrow$ 0.58 $\uparrow$ 0.68 $\downarrow$ 0.83} \\
NINCO & \multicolumn{1}{c|}{$\uparrow$ \textbf{0.72} $\uparrow$ \textbf{0.85} $\downarrow$ \textbf{0.66}} & \multicolumn{1}{c|}{$\uparrow$ 0.71 $\uparrow$ 0.82 $\downarrow$ 0.71} & \multicolumn{1}{c|}{$\uparrow$ 0.69 $\uparrow$  0.81 $\downarrow$ 0.68} & \multicolumn{1}{c|}{$\uparrow$ 0.41 $\uparrow$ 0.47 $\downarrow$ 0.98} & \multicolumn{1}{c}{$\uparrow$ 0.66 $\uparrow$ 0.78 $\downarrow$ 0.71} \\
OpenImage-O & \multicolumn{1}{c|}{$\uparrow$ \textbf{0.77} $\uparrow$ \textbf{0.91} $\downarrow$ \textbf{0.48}} & \multicolumn{1}{c|}{$\uparrow$ 0.74 $\uparrow$ 0.86 $\downarrow$ 0.61} & \multicolumn{1}{c|}{$\uparrow$ 0.71 $\uparrow$  0.84 $\downarrow$ 0.59} & \multicolumn{1}{c|}{$\uparrow$ 0.55 $\uparrow$ 0.64 $\downarrow$ 0.95} & \multicolumn{1}{c}{$\uparrow$ 0.67 $\uparrow$ 0.80 $\downarrow$ 0.65} \\
\end{tabular}}}

\subfloat[DenseNet-121]{
\resizebox{\linewidth}{!}{
\begin{tabular}{l?ccccc}
\multicolumn{1}{c?}{\multirow{2}{*}{\textbf{\begin{tabular}[c]{@{}c@{}}Unknowns\\ \end{tabular}}}} & \multicolumn{5}{c}{\textbf{$\uparrow$ AUOSCR}  \textbf{$\uparrow$ AUROC} \textbf{$\downarrow$ FPR95}} \\ \cline{2-6} 
\multicolumn{1}{c?}{} & \multicolumn{1}{c|}{\begin{tabular}[c]{@{}c@{}}\textbf{GHOST} (ours)\end{tabular}} & \multicolumn{1}{c|}{\begin{tabular}[c]{@{}c@{}}\textbf{MSP} \end{tabular}} & \multicolumn{1}{c|}{\begin{tabular}[c]{@{}c@{}}\textbf{MaxLogit} \end{tabular}} & \multicolumn{1}{c|}{\begin{tabular}[c]{@{}c@{}}\textbf{NNGuide} \end{tabular}} & \multicolumn{1}{c}{\begin{tabular}[c]{@{}c@{}}\textbf{Energy} \end{tabular}} \\ \noalign{\hrule height 1.5pt}
21K-P \emph{Easy} & \multicolumn{1}{c|}{$\uparrow$ 0.61 $\uparrow$ 0.77 $\downarrow$ 0.75} & \multicolumn{1}{c|}{$\uparrow$ 0.63 $\uparrow$ 0.77 $\downarrow$ 0.76} & \multicolumn{1}{c|}{$\uparrow$ \textbf{0.64} $\uparrow$  \textbf{0.79} $\downarrow$ \textbf{0.74}} & \multicolumn{1}{c|}{$\uparrow$ 0.17 $\uparrow$ 0.27 $\downarrow$ 0.99} & \multicolumn{1}{c}{$\uparrow$ 0.63 $\uparrow$ \textbf{0.79} $\downarrow$ \textbf{0.74}} \\
21K-P \emph{Hard} & \multicolumn{1}{c|}{$\uparrow$ 0.58 $\uparrow$ \textbf{0.72} $\downarrow$ \textbf{0.85}} & \multicolumn{1}{c|}{$\uparrow$ \textbf{0.60} $\uparrow$ 0.71 $\downarrow$ 0.87} & \multicolumn{1}{c|}{$\uparrow$ 0.57 $\uparrow$  0.70 $\downarrow$ 0.87} & \multicolumn{1}{c|}{$\uparrow$ 0.20 $\uparrow$ 0.32 $\downarrow$ 1.00} & \multicolumn{1}{c}{$\uparrow$ 0.56 $\uparrow$ 0.69 $\downarrow$ 0.87}  \\
NINCO & \multicolumn{1}{c|}{$\uparrow$ \textbf{0.65} $\uparrow$ \textbf{0.82} $\downarrow$ \textbf{0.74}} & \multicolumn{1}{c|}{$\uparrow$ \textbf{0.65} $\uparrow$ 0.79 $\downarrow$ 0.77} & \multicolumn{1}{c|}{$\uparrow$ 0.64 $\uparrow$  0.79 $\downarrow$ 0.77} & \multicolumn{1}{c|}{$\uparrow$ 0.16 $\uparrow$ 0.25 $\downarrow$ 1.00} & \multicolumn{1}{c}{$\uparrow$ 0.63 $\uparrow$ 0.79 $\downarrow$ 0.77} \\
OpenImage-O & \multicolumn{1}{c|}{$\uparrow$ \textbf{0.69} $\uparrow$ \textbf{0.89} $\downarrow$ \textbf{0.56}} & \multicolumn{1}{c|}{$\uparrow$ 0.68 $\uparrow$ 0.84 $\downarrow$ 0.67} & \multicolumn{1}{c|}{$\uparrow$ 0.69 $\uparrow$  0.88 $\downarrow$ 0.58} & \multicolumn{1}{c|}{$\uparrow$ 0.09 $\uparrow$ 0.15 $\downarrow$ 1.00} & \multicolumn{1}{c}{$\uparrow$ 0.69 $\uparrow$ 0.88 $\downarrow$ 0.57}  \\
\end{tabular}}}

\subfloat[MAE-H]{
\resizebox{\linewidth}{!}{
\begin{tabular}{l?cccccccc}
\multicolumn{1}{c?}{\multirow{2}{*}{\textbf{\begin{tabular}[c]{@{}c@{}}Unknowns\\ \end{tabular}}}} & \multicolumn{6}{c}{\textbf{$\uparrow$ AUOSCR}  \textbf{$\uparrow$ AUROC} \textbf{$\downarrow$ FPR95}} \\ \cline{2-9} 
\multicolumn{1}{c?}{} & \multicolumn{1}{c|}{\begin{tabular}[c]{@{}c@{}}\textbf{GHOST} (ours)\end{tabular}} & \multicolumn{1}{c|}{\begin{tabular}[c]{@{}c@{}}\textbf{MSP} \end{tabular}} & \multicolumn{1}{c|}{\begin{tabular}[c]{@{}c@{}}\textbf{MaxLogit} \end{tabular}} & \multicolumn{1}{c|}{\begin{tabular}[c]{@{}c@{}}\textbf{NNGuide} \end{tabular}} & \multicolumn{1}{c|}{\begin{tabular}[c]{@{}c@{}}\textbf{Energy} \end{tabular}} & \multicolumn{1}{c|}{\begin{tabular}[c]{@{}c@{}}\textbf{SCALE} \end{tabular}} & \multicolumn{1}{c|}{\begin{tabular}[c]{@{}c@{}}\textbf{REACT} \end{tabular}} & \multicolumn{1}{c}{\begin{tabular}[c]{@{}c@{}}\textbf{KNN} \end{tabular}}\\

\noalign{\hrule height 1.5pt}

21K-P \emph{Easy} & \multicolumn{1}{c|}{$\uparrow$ 0.75 $\uparrow$ 0.84 $\downarrow$ 0.58} & \multicolumn{1}{c|}{$\uparrow$ .0.72 $\uparrow$ .0.79 $\downarrow$ 0.65} & \multicolumn{1}{c|}{$\uparrow$ 0.67 $\uparrow$  0.75 $\downarrow$ 0.63} & \multicolumn{1}{c|}{$\uparrow$ 0.62 $\uparrow$ 0.69 $\downarrow$ 0.80} & \multicolumn{1}{c|}{$\uparrow$ 0.63 $\uparrow$ 0.70 $\downarrow$ 0.70} & \multicolumn{1}{c|}{$\uparrow$ 0.59 $\uparrow$ 0.66 $\downarrow$ 0.73} & \multicolumn{1}{c|}{$\uparrow$ 0.75 $\uparrow$ 0.84 $\downarrow$ 0.55} & \multicolumn{1}{c}{$\uparrow$ \textbf{0.77} $\uparrow$ \textbf{0.86} $\downarrow$ \textbf{0.52}}\\

21K-P \emph{Hard} & \multicolumn{1}{c|}{$\uparrow$ \textbf{0.73} $\uparrow$ \textbf{0.81} $\downarrow$ \textbf{0.62}} & \multicolumn{1}{c|}{$\uparrow$ .0.69 $\uparrow$ .0.75 $\downarrow$ 0.75} & \multicolumn{1}{c|}{$\uparrow$ 0.65 $\uparrow$  0.71 $\downarrow$ 0.74} & \multicolumn{1}{c|}{$\uparrow$ 0.47 $\uparrow$ 0.52 $\downarrow$ 0.89} & \multicolumn{1}{c|}{$\uparrow$ 0.61 $\uparrow$ 0.68 $\downarrow$ 0.79} & \multicolumn{1}{c|}{$\uparrow$ 0.59 $\uparrow$ 0.66 $\downarrow$ 0.80} & \multicolumn{1}{c|}{$\uparrow$ 0.68 $\uparrow$ 0.75 $\downarrow$ 0.73} & \multicolumn{1}{c}{$\uparrow$ 0.63 $\uparrow$ 0.70 $\downarrow$ 0.80} \\

NINCO & \multicolumn{1}{c|}{$\uparrow$ \textbf{0.81} $\uparrow$ \textbf{0.91} $\downarrow$ \textbf{0.47}} & \multicolumn{1}{c|}{$\uparrow$ .0.76 $\uparrow$ .0.83 $\downarrow$ 0.65} & \multicolumn{1}{c|}{$\uparrow$ 0.71 $\uparrow$  0.79 $\downarrow$ 0.62} & \multicolumn{1}{c|}{$\uparrow$ 0.49 $\uparrow$ 0.55 $\downarrow$ 0.88} & \multicolumn{1}{c|}{$\uparrow$ 0.66 $\uparrow$ 0.74 $\downarrow$ 0.69} & \multicolumn{1}{c|}{$\uparrow$ 0.63 $\uparrow$ 0.71 $\downarrow$ 0.71} & \multicolumn{1}{c|}{$\uparrow$ 0.77 $\uparrow$ 0.86 $\downarrow$ 0.57} & \multicolumn{1}{c}{$\uparrow$ 0.75 $\uparrow$ 0.84 $\downarrow$ 0.64} \\

OpenImage-O & \multicolumn{1}{c|}{$\uparrow$ \textbf{0.84} $\uparrow$ \textbf{0.95} $\downarrow$ \textbf{0.26}} & \multicolumn{1}{c|}{$\uparrow$ .0.78 $\uparrow$ .0.87 $\downarrow$ 0.52} & \multicolumn{1}{c|}{$\uparrow$ 0.73 $\uparrow$  0.82 $\downarrow$ 0.49} & \multicolumn{1}{c|}{$\uparrow$ 0.68 $\uparrow$ 0.77 $\downarrow$ 0.64} & \multicolumn{1}{c|}{$\uparrow$ 0.68 $\uparrow$ 0.77 $\downarrow$ 0.56} & \multicolumn{1}{c|}{$\uparrow$ 0.65 $\uparrow$ 0.73 $\downarrow$ 0.62} & \multicolumn{1}{c|}{$\uparrow$ 0.83 $\uparrow$ 0.93 $\downarrow$ 0.34} & \multicolumn{1}{c}{$\uparrow$ 0.83 $\uparrow$ 0.94 $\downarrow$ 0.30}\\
\end{tabular}}}

\subfloat[ConvNextV2-H]{
\resizebox{\linewidth}{!}{
\begin{tabular}{l?cccccccc}
\multicolumn{1}{c?}{\multirow{2}{*}{\textbf{\begin{tabular}[c]{@{}c@{}}Unknowns\\ \end{tabular}}}} & \multicolumn{6}{c}{\textbf{$\uparrow$ AUOSCR}  \textbf{$\uparrow$ AUROC} \textbf{$\downarrow$ FPR95}} \\ \cline{2-9} 
\multicolumn{1}{c?}{} & \multicolumn{1}{c|}{\begin{tabular}[c]{@{}c@{}}\textbf{GHOST} (ours)\end{tabular}} & \multicolumn{1}{c|}{\begin{tabular}[c]{@{}c@{}}\textbf{MSP} \end{tabular}} & \multicolumn{1}{c|}{\begin{tabular}[c]{@{}c@{}}\textbf{MaxLogit} \end{tabular}} & \multicolumn{1}{c|}{\begin{tabular}[c]{@{}c@{}}\textbf{NNGuide} \end{tabular}} & \multicolumn{1}{c|}{\begin{tabular}[c]{@{}c@{}}\textbf{Energy} \end{tabular}} & \multicolumn{1}{c|}{\begin{tabular}[c]{@{}c@{}}\textbf{SCALE} \end{tabular}} & \multicolumn{1}{c|}{\begin{tabular}[c]{@{}c@{}}\textbf{REACT} \end{tabular}} & \multicolumn{1}{c}{\begin{tabular}[c]{@{}c@{}}\textbf{KNN} \end{tabular}}\ \\
\noalign{\hrule height 1.5pt}
21K-P \emph{Easy} & \multicolumn{1}{c|}{$\uparrow$ \textbf{0.74} $\uparrow$ 0.83 $\downarrow$ 0.60} & \multicolumn{1}{c|}{$\uparrow$ 0.72 $\uparrow$ 0.79 $\downarrow$ 0.65} & \multicolumn{1}{c|}{$\uparrow$ 0.68 $\uparrow$  0.75 $\downarrow$ 0.64} & \multicolumn{1}{c|}{$\uparrow$ 0.70 $\uparrow$ 0.79 $\downarrow$ 0.70} & \multicolumn{1}{c|}{$\uparrow$ 0.64 $\uparrow$ 0.72 $\downarrow$ 0.69} & \multicolumn{1}{c|}{$\uparrow$ 0.58 $\uparrow$ 0.66 $\downarrow$ 0.81} & \multicolumn{1}{c|}{$\uparrow$ \textbf{0.74} $\uparrow$ 0.83 $\downarrow$ 0.56} & \multicolumn{1}{c}{$\uparrow$ \textbf{0.74} $\uparrow$ \textbf{0.84} $\downarrow$ \textbf{0.53}} \\

21K-P \emph{Hard} & \multicolumn{1}{c|}{$\uparrow$ \textbf{0.72} $\uparrow$ \textbf{0.80} $\downarrow$ \textbf{0.65}} & \multicolumn{1}{c|}{$\uparrow$ 0.68 $\uparrow$ 0.74 $\downarrow$ 0.76} & \multicolumn{1}{c|}{$\uparrow$ 0.65 $\uparrow$  0.72 $\downarrow$ 0.74} & \multicolumn{1}{c|}{$\uparrow$ 0.60 $\uparrow$ 0.67 $\downarrow$ 0.83} & \multicolumn{1}{c|}{$\uparrow$ 0.62 $\uparrow$ 0.69 $\downarrow$ 0.78} & \multicolumn{1}{c|}{$\uparrow$ 0.55 $\uparrow$ 0.62 $\downarrow$ 0.87} & \multicolumn{1}{c|}{$\uparrow$ 0.69 $\uparrow$ 0.75 $\downarrow$ 0.72} & \multicolumn{1}{c}{$\uparrow$ 0.58 $\uparrow$ 0.65 $\downarrow$ 0.81} \\

NINCO & \multicolumn{1}{c|}{$\uparrow$ \textbf{0.79} $\uparrow$ \textbf{0.89} $\downarrow$ \textbf{0.50}} & \multicolumn{1}{c|}{$\uparrow$ 0.75 $\uparrow$ 0.83 $\downarrow$ 0.64} & \multicolumn{1}{c|}{$\uparrow$ 0.73 $\uparrow$  0.82 $\downarrow$ 0.60} & \multicolumn{1}{c|}{$\uparrow$ 0.66 $\uparrow$ 0.74 $\downarrow$ 0.78} & \multicolumn{1}{c|}{$\uparrow$ 0.71 $\uparrow$ 0.80 $\downarrow$ 0.63} & \multicolumn{1}{c|}{$\uparrow$ 0.55 $\uparrow$ 0.62 $\downarrow$ 0.86} & \multicolumn{1}{c|}{$\uparrow$ 0.78 $\uparrow$ 0.87 $\downarrow$ 0.53} & \multicolumn{1}{c}{$\uparrow$ 0.70 $\uparrow$ 0.80 $\downarrow$ 0.67} \\

OpenImage-O & \multicolumn{1}{c|}{$\uparrow$ \textbf{0.83} $\uparrow$ \textbf{0.94} $\downarrow$ \textbf{0.32}} & \multicolumn{1}{c|}{$\uparrow$ 0.79 $\uparrow$ 0.88 $\downarrow$ 0.49} & \multicolumn{1}{c|}{$\uparrow$ 0.77 $\uparrow$  0.87 $\downarrow$ 0.44} & \multicolumn{1}{c|}{$\uparrow$ 0.74 $\uparrow$ 0.83 $\downarrow$ 0.64} & \multicolumn{1}{c|}{$\uparrow$ 0.75 $\uparrow$ 0.85 $\downarrow$ 0.47} & \multicolumn{1}{c|}{$\uparrow$ 0.68 $\uparrow$ 0.77 $\downarrow$ 0.69} & \multicolumn{1}{c|}{$\uparrow$ 0.82 $\uparrow$ 0.92 $\downarrow$ 0.34} & \multicolumn{1}{c}{$\uparrow$ 0.80 $\uparrow$ 0.92 $\downarrow$ 0.38} \\

\end{tabular}}}
\end{table*}

\begin{table*}[t!]
\centering
\small
\Caption[tab:coeff-f95c]{Coefficient of Variance and F@C95 Results}{This table includes the unfairness measure ${\cal V}_{\mathrm{CCR}}$ coefficients ($\downarrow$) of all methods, as well as the corresponding minimum FPR at 95\% of closed set accuracy ($\downarrow$). Each is computed on three pre-trained networks at 10\% FPR and evaluated on various unknown datasets.}
\subfloat[Swin-T Coefficient of Variance]{
\resizebox{.48\linewidth}{!}{
\begin{tabular}{@{}l?c|c|c|c|cc@{}}
\textbf{Unknowns} & \textbf{GHOST} & \textbf{MSP} & \textbf{MaxLogit} & \textbf{NNGuide} & \textbf{Energy}\\ \noalign{\hrule height 1.5pt}
21K-P \emph{Easy}  & \textbf{0.37}  & 0.52    & 0.59      & 1.13   & 0.69 \\
21K-P \emph{Hard}  & \textbf{0.39}  & 0.56    & 0.55      & 2.87   & 0.61 \\
NINCO              & \textbf{0.26}  & 0.45        & 0.36          & 2.42       & 0.40  \\
OpenImage-O        & \textbf{0.22}  & 0.39  & 0.33    & 1.25 & 0.39  
\end{tabular}}}\quad
\subfloat[Swin-T F@C95]{
\resizebox{.48\linewidth}{!}{
\begin{tabular}{@{}l?c|c|c|c|cc@{}}
\textbf{Unknowns}  & \textbf{GHOST} & \textbf{MSP}         & \textbf{MaxLogit} & \textbf{NNGuide} & \textbf{Energy} \\ \noalign{\hrule height 1.5pt}
21K-P \emph{Easy}  & 0.57  & \textbf{0.53}    & 0.56      & 0.92   & 0.68  \\
21K-P \emph{Hard}  & \textbf{0.63}  & 0.67    & 0.68      & 0.97   & 0.75  \\
NINCO              & \textbf{0.49}  & 0.50        & 0.50          & 0.97       & 0.60  \\
OpenImage-O        & \textbf{0.31}  & 0.39  & 0.40    & 0.94 & 0.54  
\end{tabular}}}

\subfloat[DenseNet-121 Coefficient of Variance]{
\resizebox{.48\linewidth}{!}{
\begin{tabular}{@{}l?c|c|c|c|cc@{}}
\textbf{Unknowns} & \textbf{GHOST} & \textbf{MSP} & \textbf{MaxLogit} & \textbf{NNGuide} & \textbf{Energy} &  \textbf{}\\ \noalign{\hrule height 1.5pt}
21K-P \emph{Easy}  & \textbf{0.38}  & 0.59    & 0.51      & 2.91   & 0.51  \\
21K-P \emph{Hard}  & \textbf{0.42}  & 0.71    & 0.65      & 2.21   & 0.66  \\
NINCO              & \textbf{0.32}  & 0.51        & 0.49          & 3.72       & 0.50   \\
OpenImage-O        & \textbf{0.28}  & 0.44  & 0.35    & 8.48 & 0.35 
\end{tabular}}}\quad
\subfloat[DenseNet-121 F@C95]{
\resizebox{.48\linewidth}{!}{
\begin{tabular}{@{}l?c|c|c|c|cc@{}}
\textbf{Unknowns}  & \textbf{GHOST} & \textbf{MSP}         & \textbf{MaxLogit} & \textbf{NNGuide} & \textbf{Energy} & \textbf{} \\ \noalign{\hrule height 1.5pt}
21K-P \emph{Easy}  & 0.59  & \textbf{0.54}    & 0.55      & 0.99   & \textbf{0.59}  \\
21K-P \emph{Hard}  & \textbf{0.71}  & 0.68    & 0.73      & 1.00   & 0.76  \\
NINCO              & \textbf{0.55}  & 0.53        & 0.58          & 1.00      & 0.62    \\
OpenImage-O        & \textbf{0.35}  & 0.41  & 0.35    & 1.00 & 0.38  
\end{tabular}}}

\subfloat[MAE-H Coefficient of Variance]{
\resizebox{.48\linewidth}{!}{
\begin{tabular}{@{}l?c|c|c|c|c|c@{}}
\textbf{Unknowns} & \textbf{GHOST} & \textbf{MSP} & \textbf{MaxLogit} & \textbf{NNGuide} & \textbf{Energy} &  \textbf{SCALE}\\ \noalign{\hrule height 1.5pt}
21K-P \emph{Easy}  & \textbf{0.32}  & 0.55    & 0.68      & 1.35   & 0.83 & 1.30 \\
21K-P \emph{Hard}  & \textbf{0.35}  & 0.60    & 0.61      & 2.28   & 0.69 & 1.09 \\
NINCO              & \textbf{0.21}  & 0.45        & 0.50         & 2.18       & 0.68 & 1.09   \\
OpenImage-O        & \textbf{0.17}  & 0.38  & 0.52    & 1.16 & 0.82  & 1.21
\end{tabular}}}\quad
\subfloat[MAE-H F@C95]{
\resizebox{.48\linewidth}{!}{
\begin{tabular}{@{}l?c|c|c|c|c|c@{}}
\textbf{Unknowns}  & \textbf{GHOST} & \textbf{MSP}         & \textbf{MaxLogit} & \textbf{NNGuide} & \textbf{Energy} & \textbf{SCALE} \\ \noalign{\hrule height 1.5pt}
21K-P \emph{Easy}  & \textbf{0.48}  & 0.53    & 0.54      & 0.76   & 0.65 & 0.69 \\
21K-P \emph{Hard}  & \textbf{0.53}  & 0.64    & 0.64      & 0.86   & 0.74 & 0.77 \\
NINCO              & \textbf{0.35}  & 0.51        & 0.51          & 0.86       & 0.62 & 0.66   \\
OpenImage-O        & \textbf{0.17}  & 0.39  & 0.39    & 0.60 & 0.50  & 0.58
\end{tabular}}}

\subfloat[ConvNeXtV2-H Coefficient of Variance]{
\resizebox{.48\linewidth}{!}{
\begin{tabular}{@{}l?c|c|c|c|c|c@{}}
\textbf{Unknowns} & \textbf{GHOST} & \textbf{MSP} & \textbf{MaxLogit} & \textbf{NNGuide} & \textbf{Energy} &  \textbf{SCALE}\\ \noalign{\hrule height 1.5pt}
21K-P \emph{Easy}  & \textbf{0.36}  & 0.59    & 0.64      & 0.92   & 0.70 & 1.39 \\
21K-P \emph{Hard}  & \textbf{0.42}  & 0.65    & 0.61      & 1.34   & 0.64 & 1.43 \\
NINCO              & \textbf{0.24}  & 0.46        & 0.40          & 1.06       & 0.41 & 1.52   \\
OpenImage-O        & \textbf{0.19}  & 0.36  & 0.34    & 0.72 & 0.34  & 0.92
\end{tabular}}}\quad
\subfloat[ConvNeXtV2-H F@C95]{
\resizebox{.48\linewidth}{!}{
\begin{tabular}{@{}l?c|c|c|c|c|c@{}}
\textbf{Unknowns}  & \textbf{GHOST} & \textbf{MSP}         & \textbf{MaxLogit} & \textbf{NNGuide} & \textbf{Energy} & \textbf{SCALE} \\ \noalign{\hrule height 1.5pt}
21K-P \emph{Easy}  & \textbf{0.49}  & 0.53    & 0.53      & 0.63   & 0.62 & 0.79 \\
21K-P \emph{Hard}  & \textbf{0.55}  & 0.64    & 0.63      & 0.79   & 0.72 & 0.85 \\
NINCO              & \textbf{0.37}  & 0.51        & 0.46          & 0.73       & 0.54 & 0.84   \\
OpenImage-O        & \textbf{0.20}  & 0.36  & 0.32    & 0.56 & 0.40 & 0.66
\end{tabular}}}
\end{table*}

\begin{figure}[t]
  \centering
  \includegraphics[width=\linewidth]{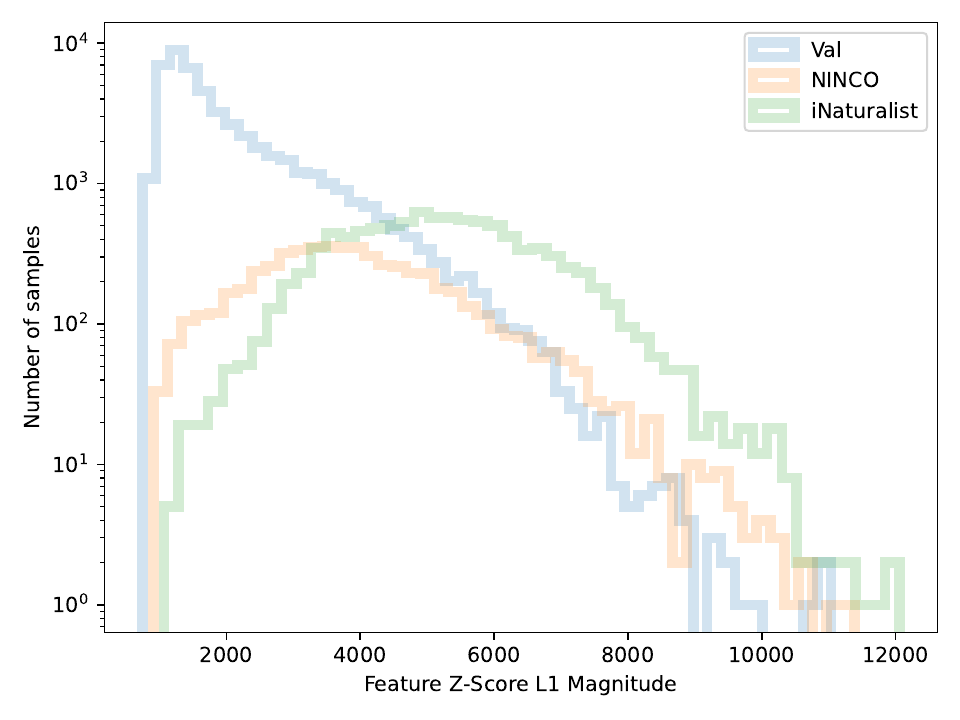}
  \Caption[fig:z-scores]{Distribution of Test-Time Z-Scores}{
    We investigate the distribution of sample z-scores according to (3) in the main paper in the validation (val) and two unknown datasets using ConvNeXtV2-H. 
    To improve visibility of failure cases, we plot the y-axis with a logarithmic axis. 
    Most val data has far lower z-scores than the peaks of NINCO and iNaturalist distributions. 
    However, as one can see, a few val samples have very large z-scores, likely due to automated data collection for ILSVRC2012 or visually distinct subclasses (e.g. baseball class contains images of baseballs and baseball games). 
    While this may cause the normality test to fail, we consider it to be only a minor issue overall, given the significant performance gains reported with GHOST. }
\end{figure}

\section{Plotted Curves}
\label{sec:plots}
We present ROC and OSCR plots on the networks from the main paper in \fig{mae-H-roc}, \fig{mae-H-oscr}, \fig{convnextv2-H-roc}, and \fig{convnextv2-H-oscr}.

\begin{figure*}[t!]
    \centering
    \subfloat[a][21K-P \emph{Easy}]{
        \includegraphics[width=0.48\textwidth]{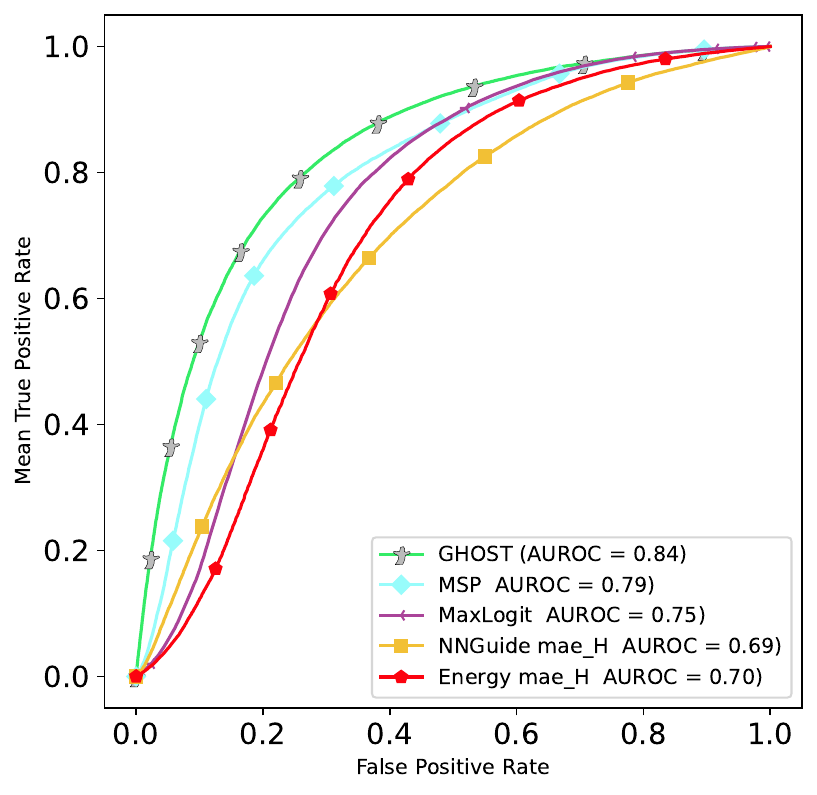}
        } 
    \subfloat[b][21K-P \emph{Hard}]{
        \includegraphics[width=0.48\textwidth]{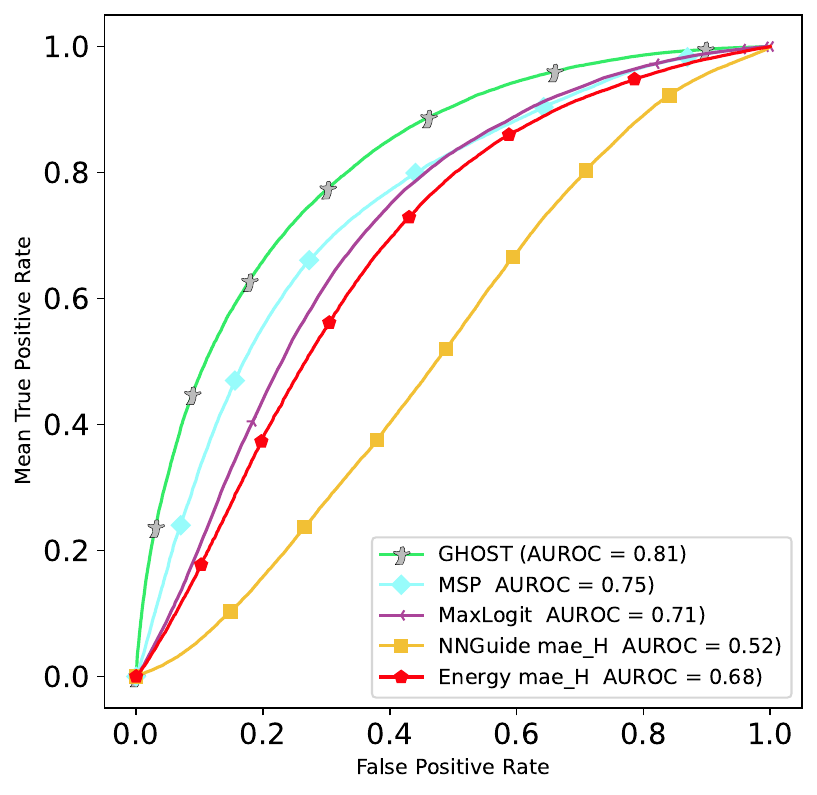}
        } \\
    \subfloat[a][NINCO]{
        \includegraphics[width=0.48\textwidth]{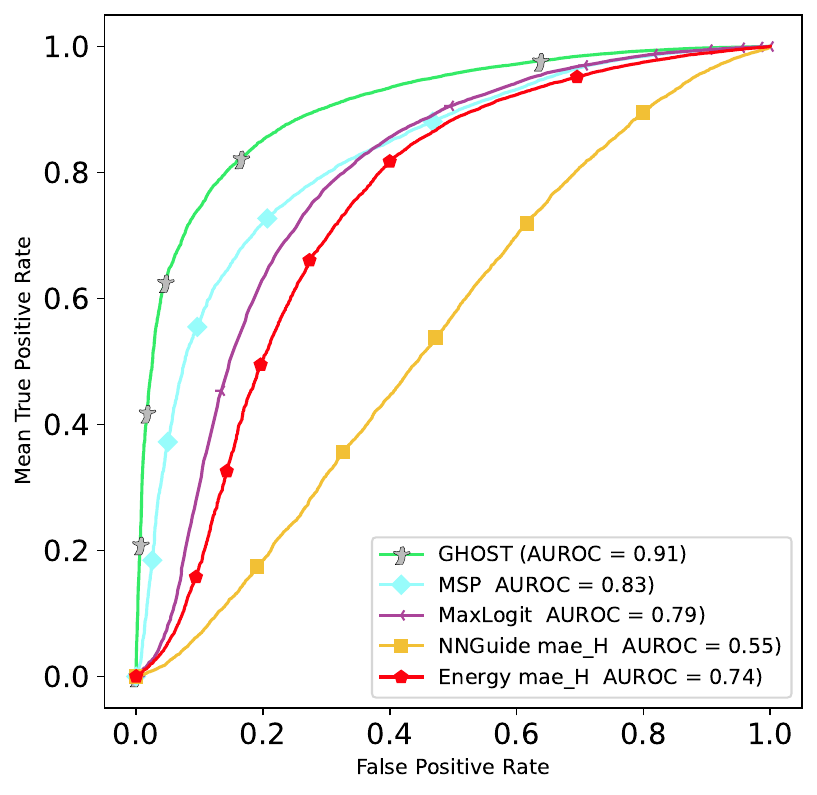}
        } 
    \subfloat[b][OpenImage-O]{
        \includegraphics[width=0.48\textwidth]{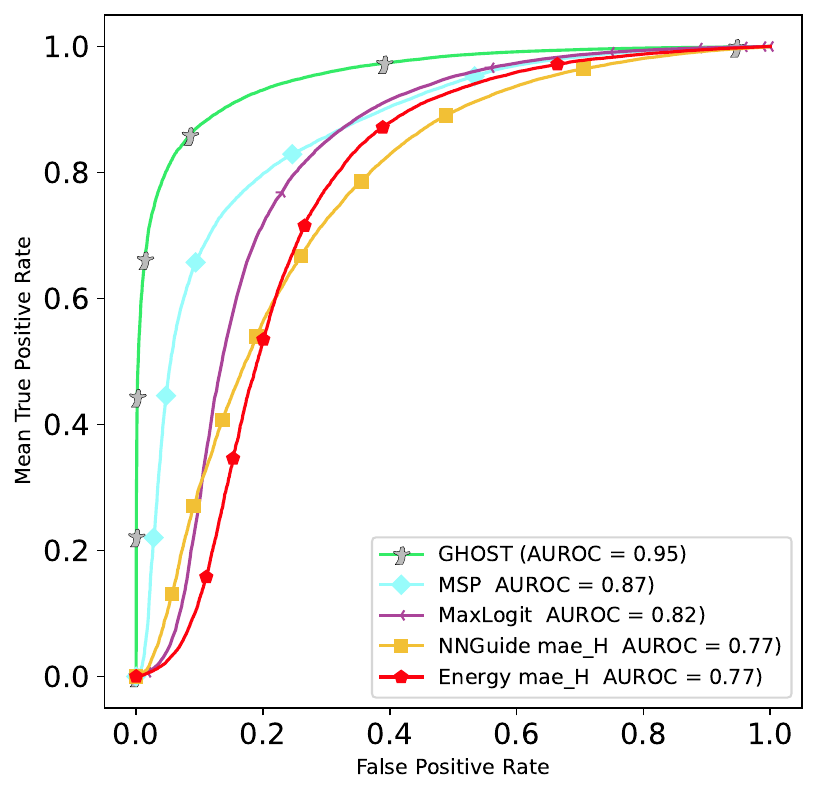}
        } \\
    \Caption[fig:mae-H-roc]{ROC Curves}{The Receiver Operating Characteristic curves of all methods on four unknown datasets from the main paper. All methods are derived from extractions from the same pre-trained architecture -- state-of-the-art MAE-H.}
\end{figure*}

\begin{figure*}[t!]
    \centering
    \subfloat[a][21K-P \emph{Easy}]{
        \includegraphics[width=0.48\textwidth]{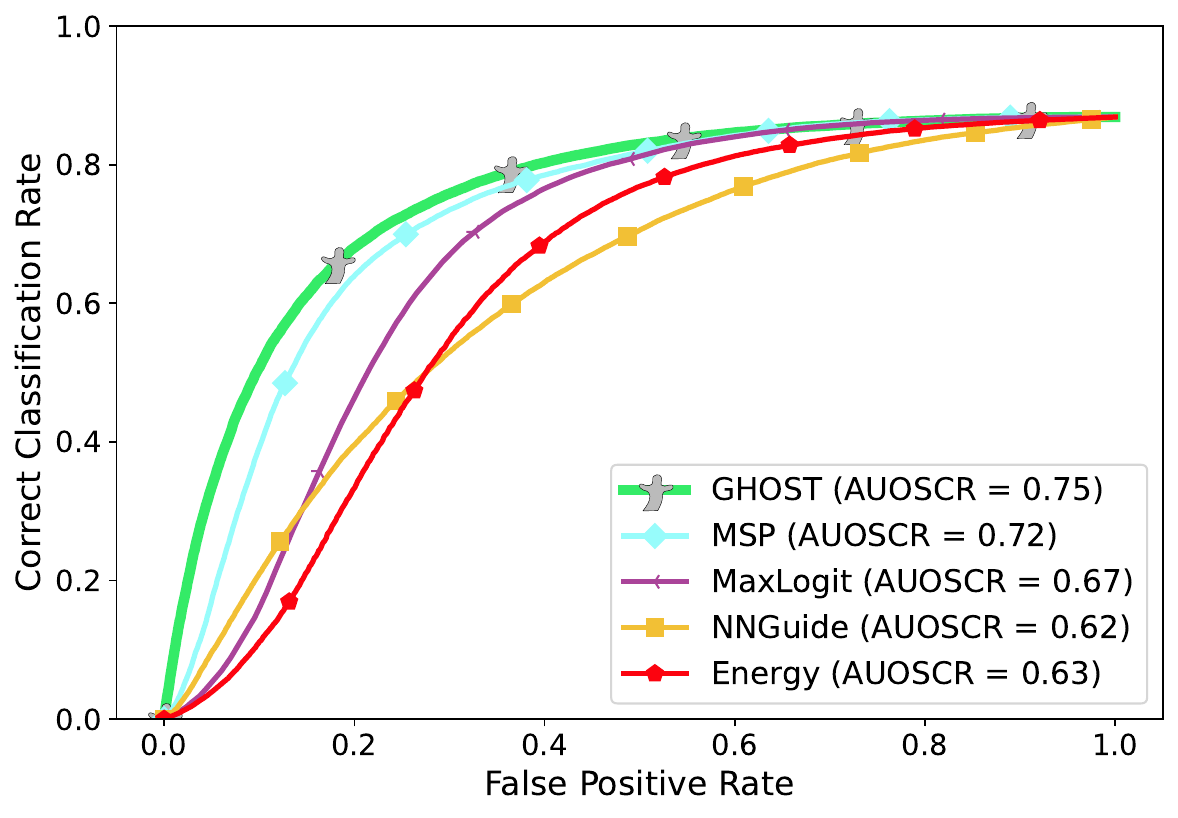}
        } 
    \subfloat[b][21K-P \emph{Hard}]{
        \includegraphics[width=0.48\textwidth]{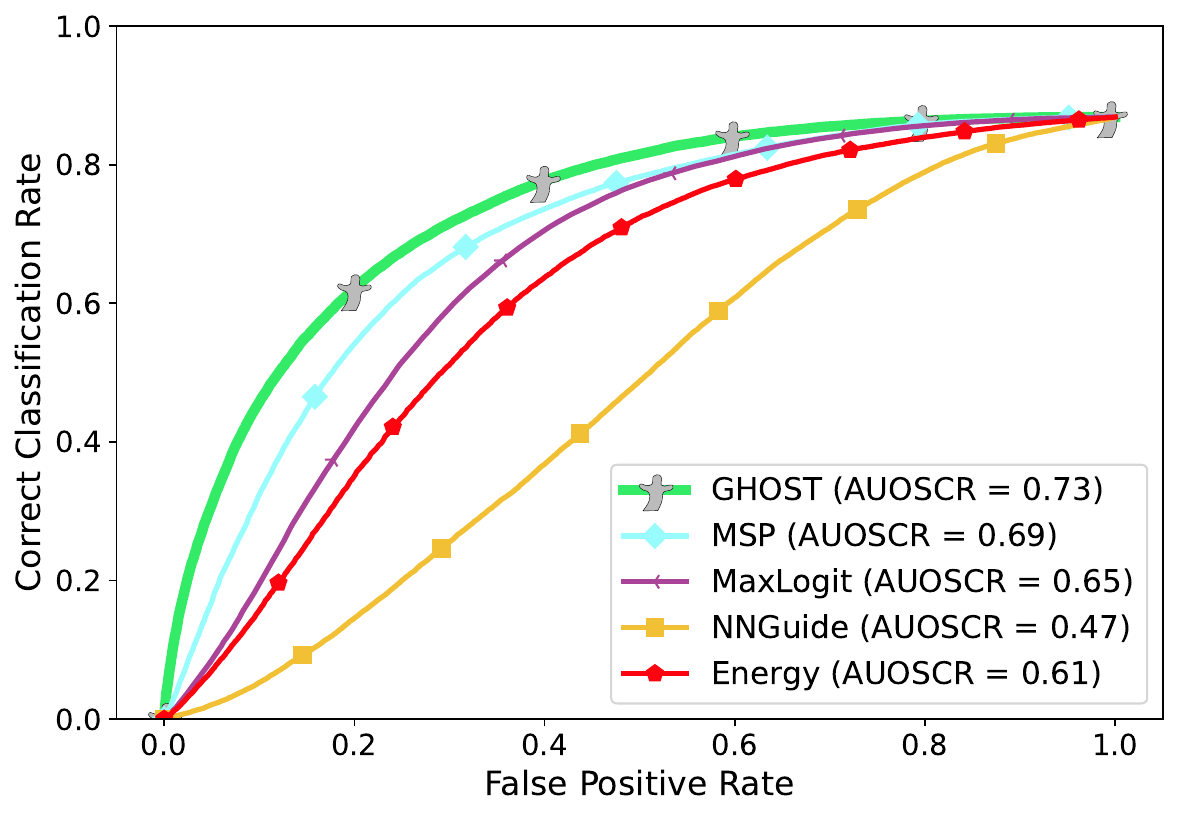}
        } \\
    \subfloat[a][NINCO]{
        \includegraphics[width=0.48\textwidth]{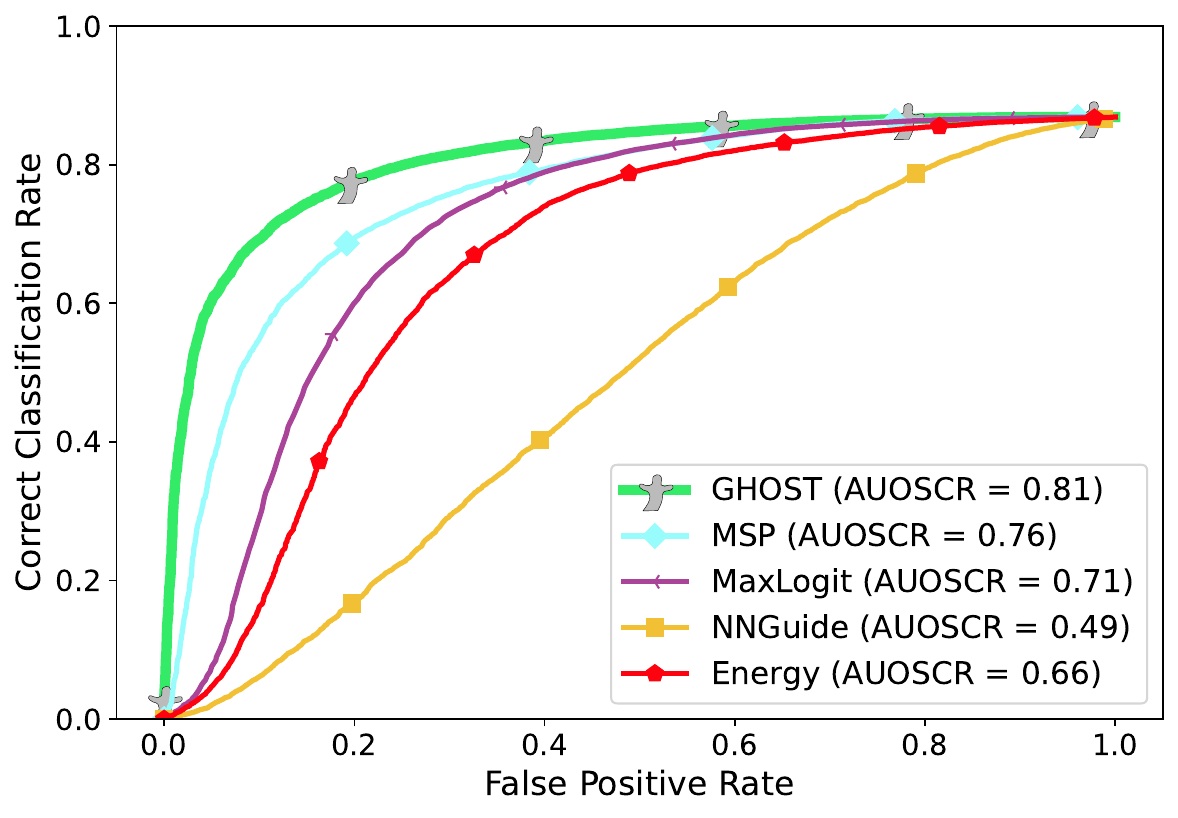}
        } 
    \subfloat[b][OpenImage-O]{
        \includegraphics[width=0.48\textwidth]{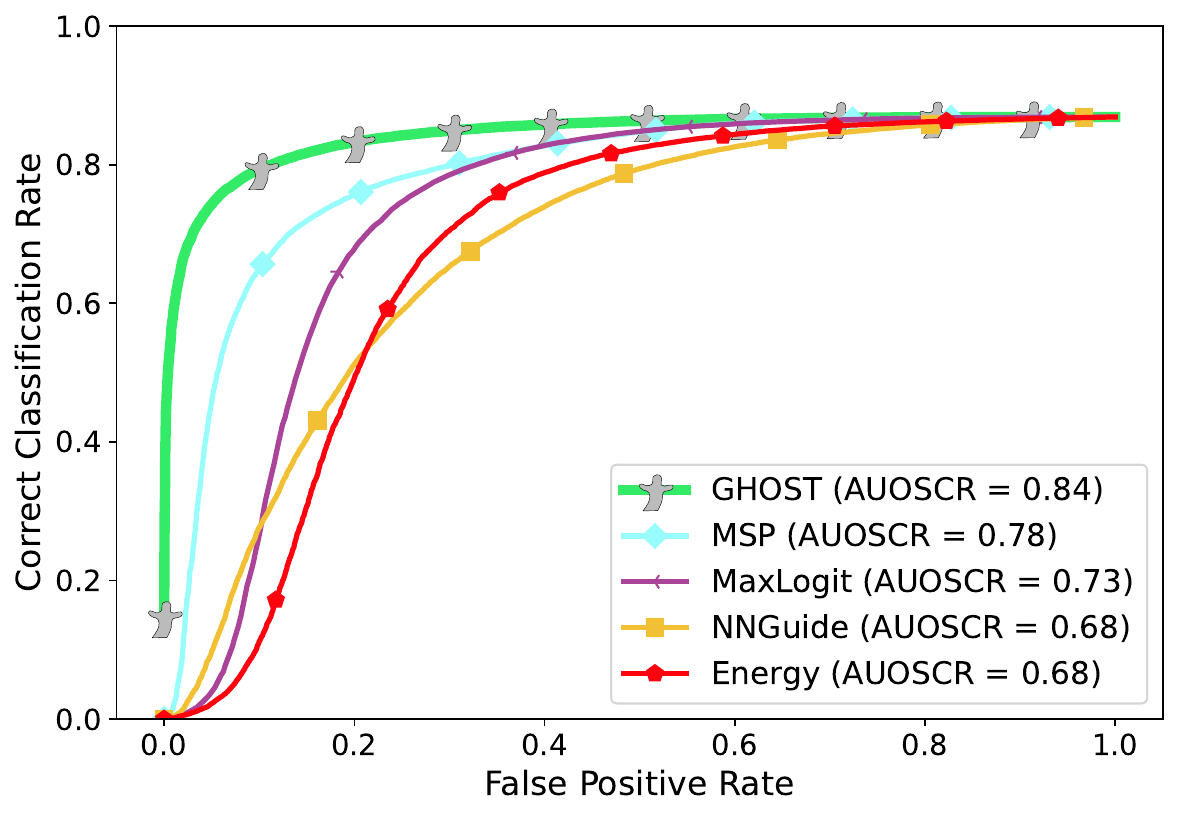}
        } \\
    \Caption[fig:mae-H-oscr]{OSCR Curves}{The Open-Set Classification Rate curves of all methods on four unknown datasets from Table 2 in the main paper. All methods are derived from extractions from the same pre-trained architecture -- state-of-the-art MAE-H.}
\end{figure*}

\begin{figure*}[t!]
    \centering
    \subfloat[a][21K-P \emph{Easy}]{
        \includegraphics[width=0.48\textwidth]{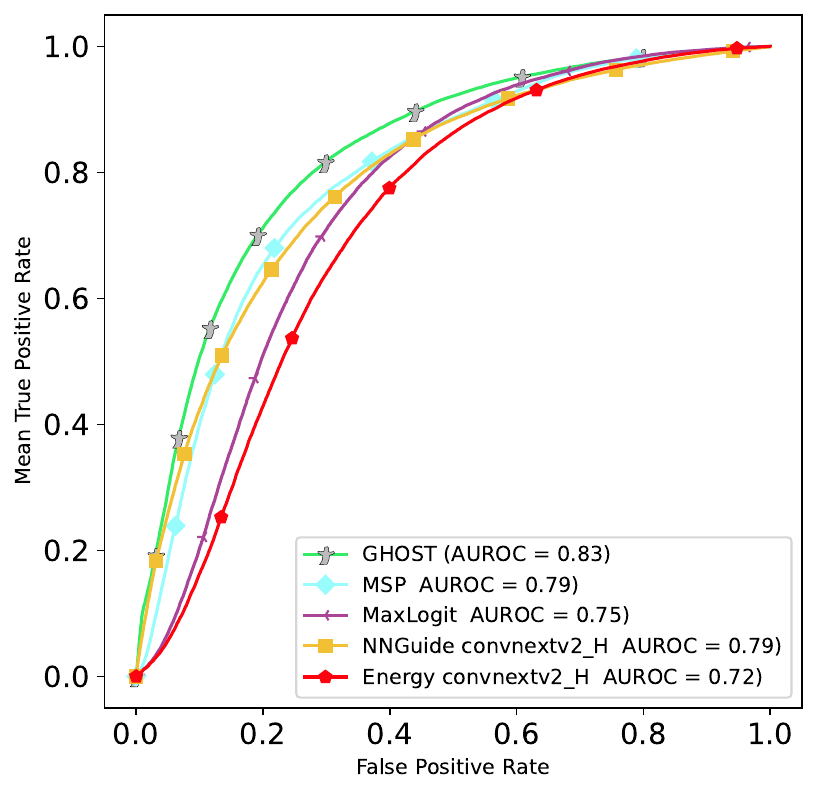}
        } 
    \subfloat[b][21K-P \emph{Hard}]{
        \includegraphics[width=0.48\textwidth]{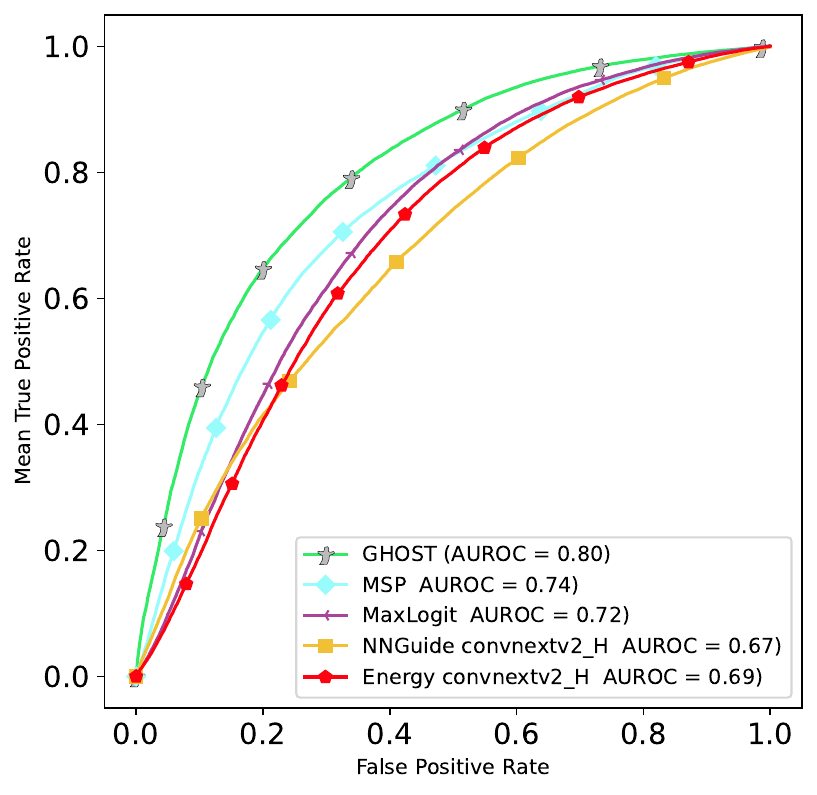}
        } \\
    \subfloat[a][NINCO]{
        \includegraphics[width=0.48\textwidth]{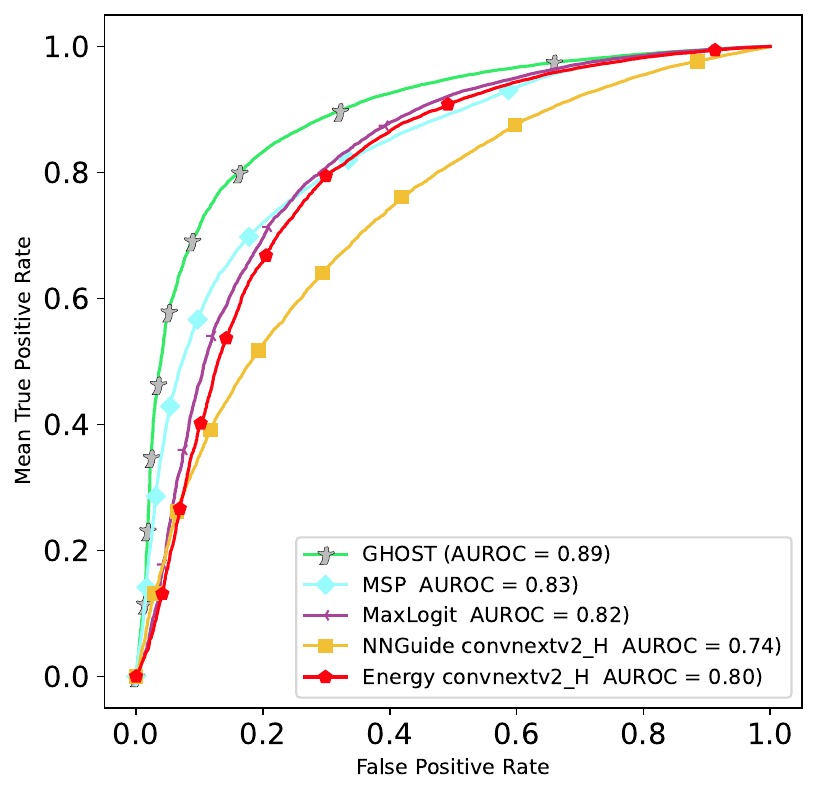}
        } 
    \subfloat[b][OpenImage-O]{
        \includegraphics[width=0.48\textwidth]{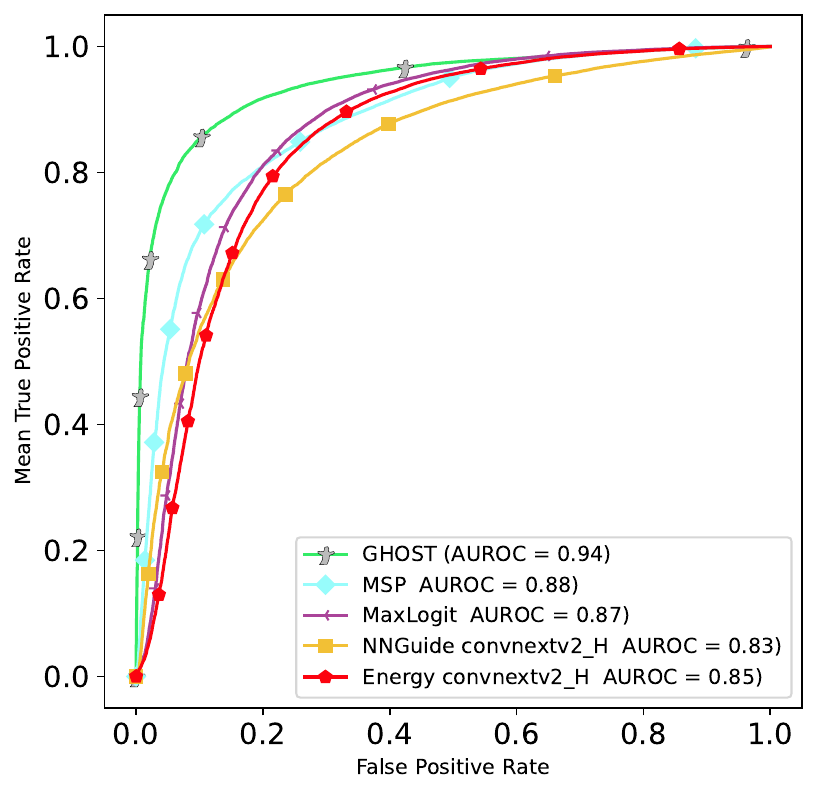}
        } \\
    \Caption[fig:convnextv2-H-roc]{ROC Curves}{The Receiver Operating Characteristic curves of all methods on four unknown datasets from the main paper. All methods are derived from extractions from the same pre-trained architecture -- state-of-the-art ConvNeXtV2-H.}
\end{figure*}

\begin{figure*}[t!]
    \centering
    \subfloat[a][21K-P \emph{Easy}]{
        \includegraphics[width=0.48\textwidth]{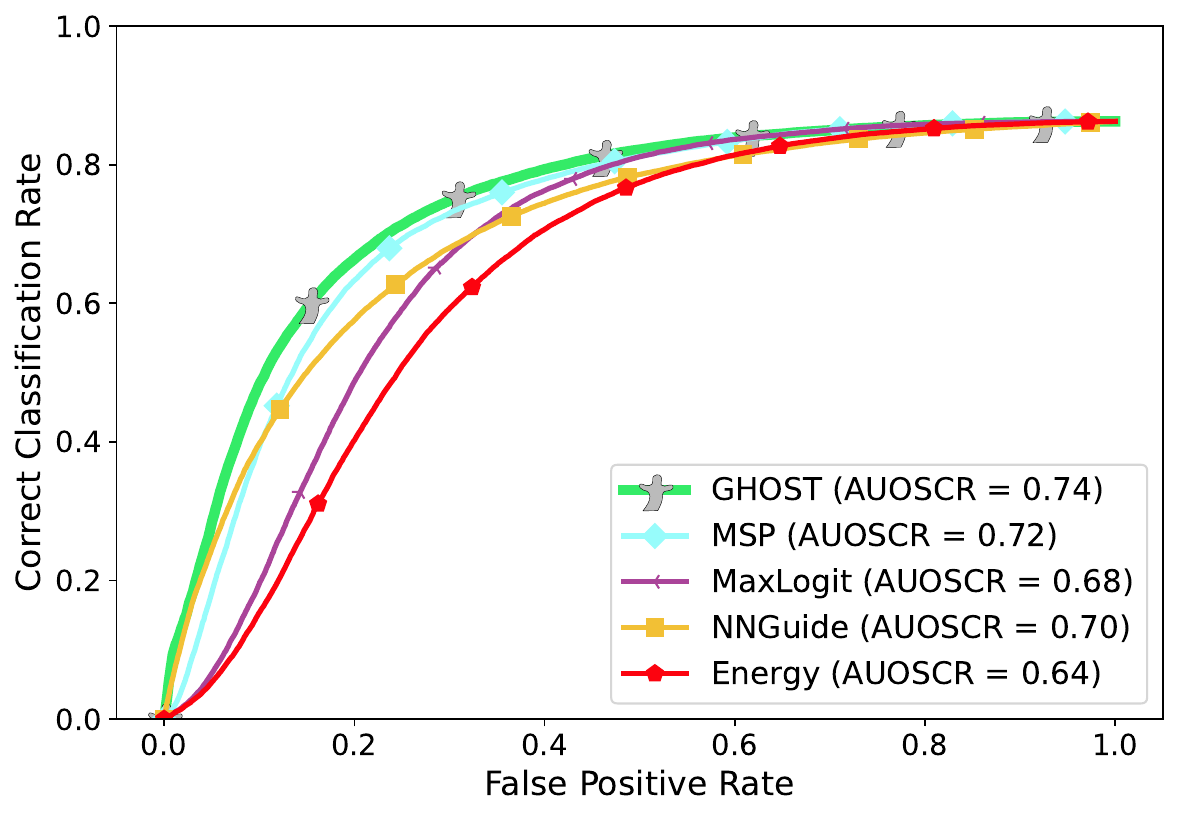}
        } 
    \subfloat[b][21K-P \emph{Hard}]{
        \includegraphics[width=0.48\textwidth]{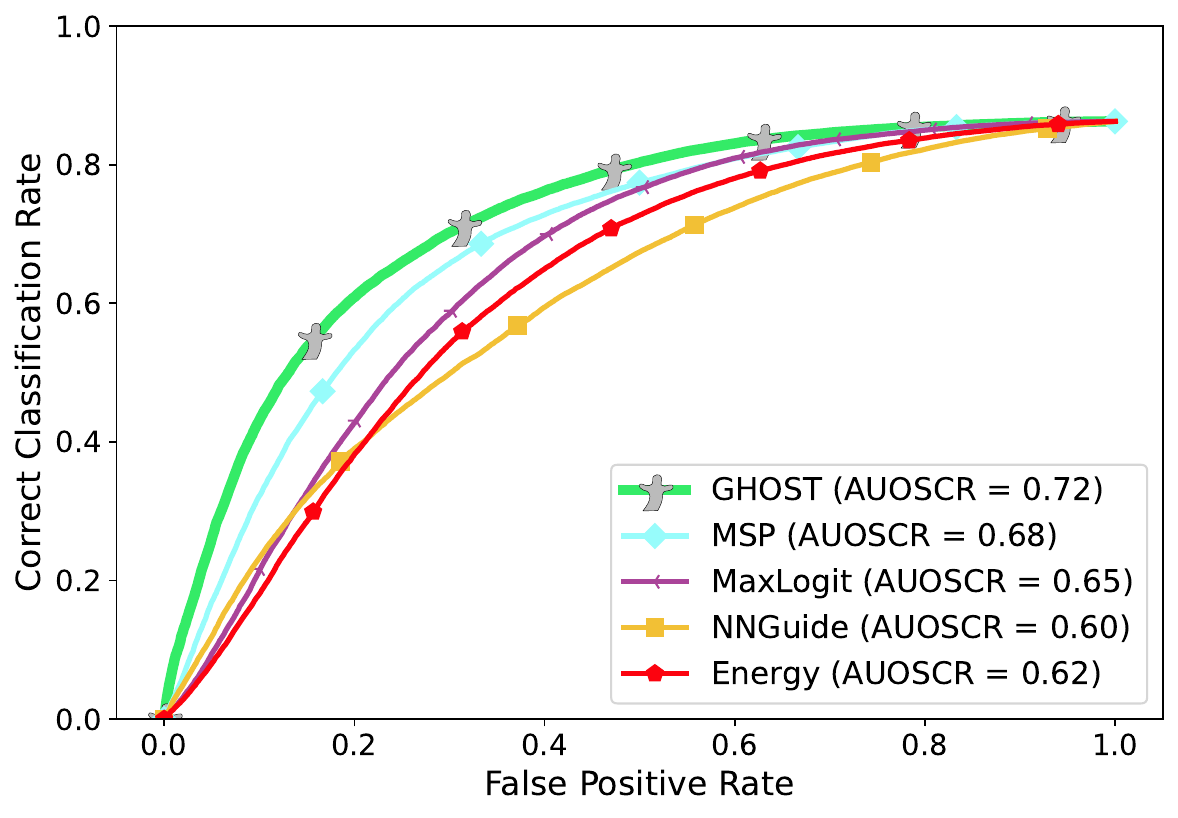}
        } \\
    \subfloat[a][NINCO]{
        \includegraphics[width=0.48\textwidth]{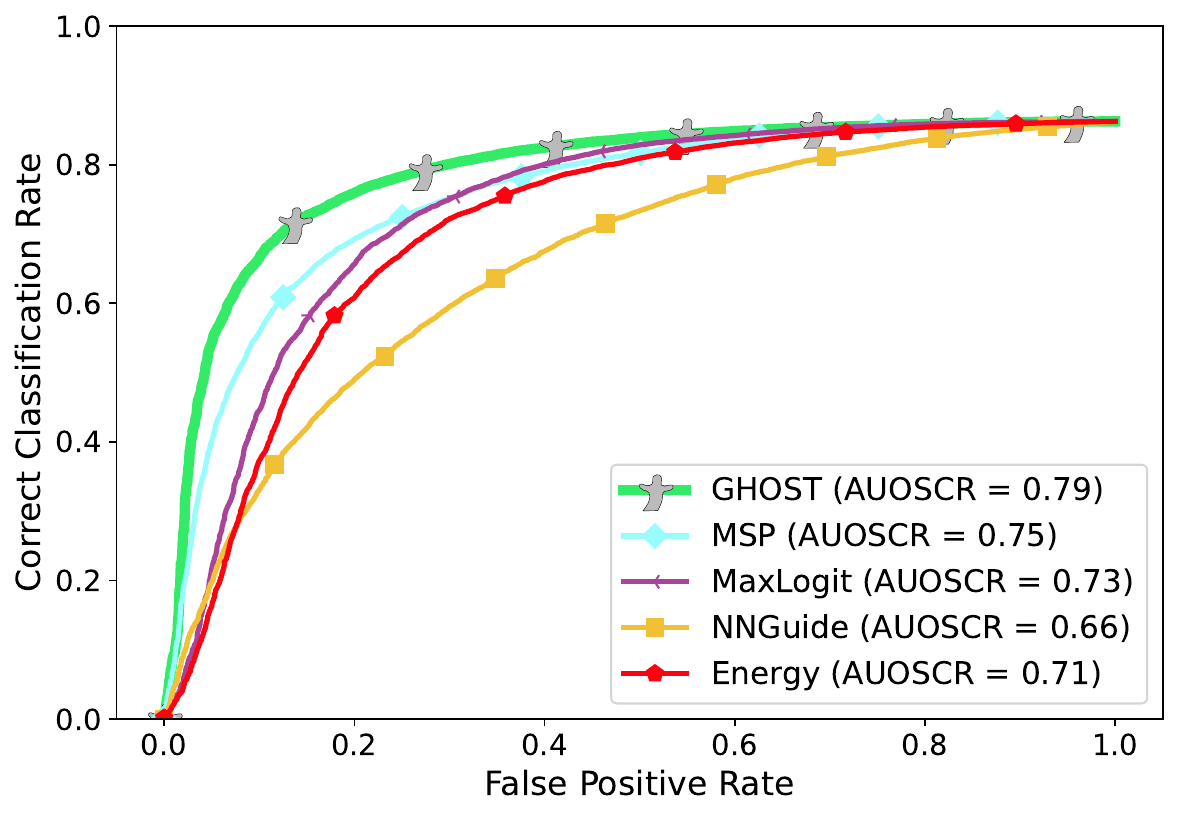}
        } 
    \subfloat[b][OpenImage-O]{
        \includegraphics[width=0.48\textwidth]{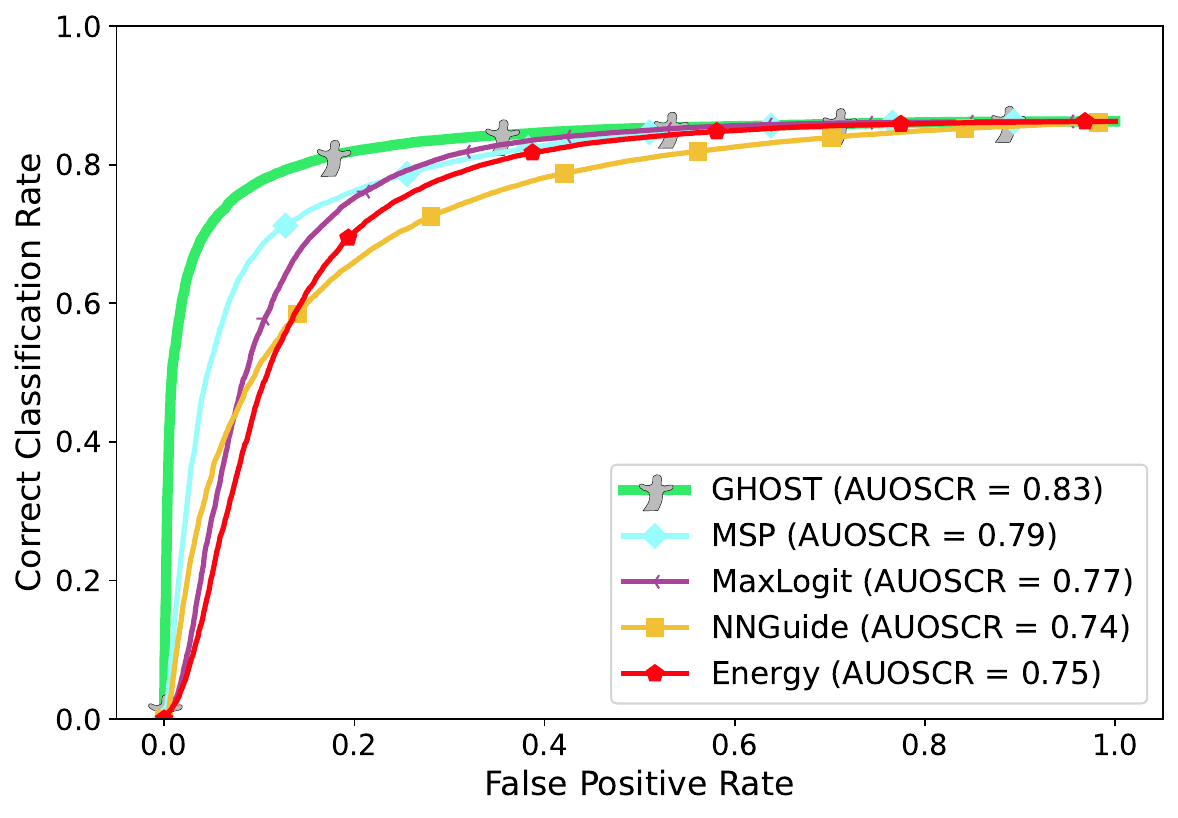}
        } \\
    \Caption[fig:convnextv2-H-oscr]{OSCR Curves}{The Open-Set Classification Rate curves of all methods on four unknown datasets from Table 2 in the main paper. All methods are derived from extractions from the same pre-trained architecture -- state-of-the-art ConvNeXtV2-H.}
\end{figure*}

\clearpage

\clearpage
\bibliography{aaai25}

\begin{thebibliography}{52}
\providecommand{\natexlab}[1]{#1}

\bibitem[{Atkinson et~al.(1970)}]{atkinson1970measurement}
Atkinson, A.~B.; et~al. 1970.
\newblock On the measurement of inequality.
\newblock \emph{Journal of economic theory}, 2(3): 244--263.

\bibitem[{Bendale and Boult(2016)}]{bendale2016openmax}
Bendale, A.; and Boult, T.~E. 2016.
\newblock Towards open set deep networks.
\newblock In \emph{Conference on Computer Vision and Pattern Recognition
  (CVPR)}. IEEE.

\bibitem[{Bisgin et~al.(2024)Bisgin, Palechor, Suter, and
  G\"unther}]{bisgin2024large}
Bisgin, H.; Palechor, A.; Suter, M.; and G\"unther, M. 2024.
\newblock Large-Scale Evaluation of Open-Set Image Classification Techniques.
\newblock \emph{arxiv}.

\bibitem[{Bitterwolf, Mueller, and Hein(2023)}]{bitterwolf2023or}
Bitterwolf, J.; Mueller, M.; and Hein, M. 2023.
\newblock In or Out? Fixing ImageNet Out-of-Distribution Detection Evaluation.
\newblock In \emph{ICLR Workshop on Trustworthy and Reliable Large-Scale
  Machine Learning Models}.

\bibitem[{Cimpoi et~al.(2014)Cimpoi, Maji, Kokkinos, Mohamed, and
  Vedaldi}]{Cimpoi_2014_CVPR}
Cimpoi, M.; Maji, S.; Kokkinos, I.; Mohamed, S.; and Vedaldi, A. 2014.
\newblock Describing Textures in the Wild.
\newblock In \emph{Conference on Computer Vision and Pattern Recognition
  (CVPR)}.

\bibitem[{Cruz et~al.(2024)Cruz, Rabinowitz, G\"unther, and
  Boult}]{cruz2024oosa}
Cruz, S.; Rabinowitz, R.; G\"unther, M.; and Boult, T.~E. 2024.
\newblock Operational Open-Set Recognition and PostMax Refinement.
\newblock In \emph{European Conference on Computer Vision (ECCV)}.

\bibitem[{Deng et~al.(2009)Deng, Dong, Socher, Li, Li, and
  Fei-Fei}]{deng2009imagenet}
Deng, J.; Dong, W.; Socher, R.; Li, L.-J.; Li, K.; and Fei-Fei, L. 2009.
\newblock ImageNet: A large-scale hierarchical image database.
\newblock In \emph{Conference on Computer Vision and Pattern Recognition
  (CVPR)}. IEEE.

\bibitem[{Dhamija, G{\"u}nther, and Boult(2018)}]{dhamija2018reducing}
Dhamija, A.~R.; G{\"u}nther, M.; and Boult, T. 2018.
\newblock Reducing network agnostophobia.
\newblock In \emph{Advances in Neural Information Processing Systems
  (NeurIPS)}.

\bibitem[{Dong et~al.(2023)Dong, Bao, Zhang, Chen, Zhang, Yuan, Chen, Wen, Yu,
  and Guo}]{dong2023peco}
Dong, X.; Bao, J.; Zhang, T.; Chen, D.; Zhang, W.; Yuan, L.; Chen, D.; Wen, F.;
  Yu, N.; and Guo, B. 2023.
\newblock Peco: Perceptual codebook for bert pre-training of vision
  transformers.
\newblock In \emph{Proceedings of the AAAI Conference on Artificial
  Intelligence}, volume~37, 552--560.

\bibitem[{Formby, Smith, and Zheng(1999)}]{formby1999coefficient}
Formby, J.~P.; Smith, W.~J.; and Zheng, B. 1999.
\newblock The coefficient of variation, stochastic dominance and inequality: a
  new interpretation.
\newblock \emph{Economics Letters}, 62(3): 319--323.

\bibitem[{Ge, Demyanov, and Garnavi(2017)}]{ge2017gopenmax}
Ge, Z.; Demyanov, S.; and Garnavi, R. 2017.
\newblock Generative {OpenMax} for Multi-Class Open Set Classification.
\newblock In \emph{British Machine Vision Conference (BMVC)}.

\bibitem[{Geng, Huang, and Chen(2020)}]{geng2020recent}
Geng, C.; Huang, S.-j.; and Chen, S. 2020.
\newblock Recent advances in open set recognition: A survey.
\newblock \emph{Transactions on Pattern Analysis and Machine Intelligence
  (TPAMI)}, 43(10).

\bibitem[{He et~al.(2022)He, Chen, Xie, Li, Doll{\'a}r, and
  Girshick}]{he2022masked}
He, K.; Chen, X.; Xie, S.; Li, Y.; Doll{\'a}r, P.; and Girshick, R. 2022.
\newblock Masked autoencoders are scalable vision learners.
\newblock In \emph{Conference on Computer Vision and Pattern Recognition
  (CVPR)}.

\bibitem[{Hendrycks et~al.(2022)Hendrycks, Basart, Mazeika, Zou, Kwon,
  Mostajabi, Steinhardt, and Song}]{hendrycks2022scaling}
Hendrycks, D.; Basart, S.; Mazeika, M.; Zou, A.; Kwon, J.; Mostajabi, M.;
  Steinhardt, J.; and Song, D. 2022.
\newblock Scaling Out-of-Distribution Detection for Real-World Settings.
\newblock In \emph{International Conference on Machine Learning (ICML)}.

\bibitem[{Hendrycks and Gimpel(2017)}]{hendrycks17baseline}
Hendrycks, D.; and Gimpel, K. 2017.
\newblock A Baseline for Detecting Misclassified and Out-of-Distribution
  Examples in Neural Networks.
\newblock \emph{International Conference on Learning Representations (ICLR)}.

\bibitem[{Islam, Pan, and Foulds(2021)}]{islam2021can}
Islam, R.; Pan, S.; and Foulds, J.~R. 2021.
\newblock Can we obtain fairness for free?
\newblock In \emph{AAAI/ACM Conference on AI, Ethics, and Society}.

\bibitem[{Li et~al.(2024{\natexlab{a}})Li, Zhang, Geng, and Chen}]{li2024all}
Li, C.; Zhang, E.; Geng, C.; and Chen, S. 2024{\natexlab{a}}.
\newblock All Beings Are Equal in Open Set Recognition.
\newblock In \emph{AAAI Conference on Artificial Intelligence}, volume~38.

\bibitem[{Li et~al.(2024{\natexlab{b}})Li, Song, Gao, Zhu, and
  Shen}]{li2024prototype}
Li, H.; Song, J.; Gao, L.; Zhu, X.; and Shen, H. 2024{\natexlab{b}}.
\newblock Prototype-based aleatoric uncertainty quantification for cross-modal
  retrieval.
\newblock \emph{Advances in Neural Information Processing Systems}, 36.

\bibitem[{Li, Wu, and Su(2023)}]{li2023accurate}
Li, X.; Wu, P.; and Su, J. 2023.
\newblock Accurate fairness: Improving individual fairness without trading
  accuracy.
\newblock In \emph{AAAI Conference on Artificial Intelligence}, volume~37.

\bibitem[{Liu et~al.(2020)Liu, Wang, Owens, and Li}]{liu2020energy}
Liu, W.; Wang, X.; Owens, J.; and Li, Y. 2020.
\newblock Energy-based out-of-distribution detection.
\newblock \emph{Advances in Neural Information Processing Systems (NeurIPS)}.

\bibitem[{Lu et~al.(2020)Lu, Ma, Lu, Lu, and Ying}]{lu2020mean}
Lu, Y.; Ma, C.; Lu, Y.; Lu, J.; and Ying, L. 2020.
\newblock A mean field analysis of deep resnet and beyond: Towards provably
  optimization via overparameterization from depth.
\newblock In \emph{International Conference on Machine Learning (ICML)}.

\bibitem[{Miller et~al.(2021)Miller, Sunderhauf, Milford, and
  Dayoub}]{miller2021class}
Miller, D.; Sunderhauf, N.; Milford, M.; and Dayoub, F. 2021.
\newblock Class anchor clustering: A loss for distance-based open set
  recognition.
\newblock In \emph{Winter Conference on Applications of Computer Vision
  (WACV)}.

\bibitem[{Nadeem, Zucker, and Hanczar(2009)}]{nadeem2009accuracy}
Nadeem, M. S.~A.; Zucker, J.-D.; and Hanczar, B. 2009.
\newblock Accuracy-rejection curves (ARCs) for comparing classification methods
  with a reject option.
\newblock In \emph{Machine Learning in Systems Biology}, 65--81. PMLR.

\bibitem[{Neal et~al.(2018)Neal, Olson, Fern, Wong, and
  Li}]{neal2018counterfactual}
Neal, L.; Olson, M.; Fern, X.; Wong, W.-K.; and Li, F. 2018.
\newblock Open set learning with counterfactual images.
\newblock In \emph{European Conference on Computer Vision (ECCV)}.

\bibitem[{Park, Jung, and Teoh(2023)}]{park2023nearest}
Park, J.; Jung, Y.~G.; and Teoh, A. B.~J. 2023.
\newblock Nearest Neighbor Guidance for Out-of-Distribution Detection.
\newblock In \emph{International Conference on Computer Vision (ICCV)}.

\bibitem[{Perera et~al.(2020)Perera, Morariu, Jain, Manjunatha, Wigington,
  Ordonez, and Patel}]{perera_generative-discriminative_2020}
Perera, P.; Morariu, V.~I.; Jain, R.; Manjunatha, V.; Wigington, C.; Ordonez,
  V.; and Patel, V.~M. 2020.
\newblock Generative-Discriminative Feature Representations for Open-Set
  Recognition.
\newblock In \emph{Conference on Computer Vision and Pattern Recognition
  (CVPR)}.

\bibitem[{Ridnik et~al.(2021)Ridnik, Ben-Baruch, Noy, and Zelnik}]{ridnik2021}
Ridnik, T.; Ben-Baruch, E.; Noy, A.; and Zelnik, L. 2021.
\newblock ImageNet-21K Pretraining for the Masses.
\newblock In \emph{NeurIPS track on Datasets and Benchmarks}.

\bibitem[{Roady et~al.(2020)Roady, Hayes, Kemker, Gonzales, and
  Kanan}]{roady2020open}
Roady, R.; Hayes, T.~L.; Kemker, R.; Gonzales, A.; and Kanan, C. 2020.
\newblock Are open set classification methods effective on large-scale
  datasets?
\newblock \emph{Plos one}, 15(9): e0238302.

\bibitem[{Rudd et~al.(2017)Rudd, Jain, Scheirer, and Boult}]{rudd2017evm}
Rudd, E.~M.; Jain, L.~P.; Scheirer, W.~J.; and Boult, T.~E. 2017.
\newblock The extreme value machine.
\newblock \emph{Transactions on Pattern Analysis and Machine Intelligence
  (TPAMI)}.

\bibitem[{Russakovsky et~al.(2015)Russakovsky, Deng, Su, Krause, Satheesh, Ma,
  Huang, Karpathy, Khosla, Bernstein, Berg, and Fei-Fei}]{ILSVRC15}
Russakovsky, O.; Deng, J.; Su, H.; Krause, J.; Satheesh, S.; Ma, S.; Huang, Z.;
  Karpathy, A.; Khosla, A.; Bernstein, M.; Berg, A.~C.; and Fei-Fei, L. 2015.
\newblock {ImageNet Large Scale Visual Recognition Challenge}.
\newblock \emph{International Journal of Computer Vision (IJCV)}, 115(3):
  211--252.

\bibitem[{Scheirer et~al.(2012)Scheirer, de~Rezende~Rocha, Sapkota, and
  Boult}]{scheirer2012toward}
Scheirer, W.~J.; de~Rezende~Rocha, A.; Sapkota, A.; and Boult, T.~E. 2012.
\newblock Toward open set recognition.
\newblock \emph{Transactions on Pattern Analysis and Machine Intelligence
  (TPAMI)}, 35(7).

\bibitem[{Scheirer, Jain, and Boult(2014)}]{scheirer2014probability}
Scheirer, W.~J.; Jain, L.~P.; and Boult, T.~E. 2014.
\newblock Probability models for open set recognition.
\newblock \emph{Transactions on Pattern Analysis and Machine Intelligence
  (TPAMI)}, 36(11).

\bibitem[{Sensoy, Kaplan, and Kandemir(2018)}]{sensoy2018evidential}
Sensoy, M.; Kaplan, L.; and Kandemir, M. 2018.
\newblock Evidential deep learning to quantify classification uncertainty.
\newblock \emph{Advances in neural information processing systems}, 31.

\bibitem[{Sirignano and Spiliopoulos(2020)}]{sirignano2020mean}
Sirignano, J.; and Spiliopoulos, K. 2020.
\newblock Mean field analysis of neural networks: A central limit theorem.
\newblock \emph{Stochastic Processes and their Applications}, 130(3):
  1820--1852.

\bibitem[{Sun, Guo, and Li(2021)}]{sun2021react}
Sun, Y.; Guo, C.; and Li, Y. 2021.
\newblock React: Out-of-distribution detection with rectified activations.
\newblock \emph{Advances in Neural Information Processing Systems}, 34:
  144--157.

\bibitem[{Sun et~al.(2022)Sun, Ming, Zhu, and Li}]{sun2022out}
Sun, Y.; Ming, Y.; Zhu, X.; and Li, Y. 2022.
\newblock Out-of-distribution detection with deep nearest neighbors.
\newblock In \emph{International Conference on Machine Learning}, 20827--20840.
  PMLR.

\bibitem[{Trawi{\'n}ski et~al.(2012)Trawi{\'n}ski, Smetek, Telec, and
  Lasota}]{trawinski2012nonparametric}
Trawi{\'n}ski, B.; Smetek, M.; Telec, Z.; and Lasota, T. 2012.
\newblock Nonparametric statistical analysis for multiple comparison of machine
  learning regression algorithms.
\newblock \emph{International Journal of Applied Mathematics and Computer
  Science}, 22(4): 867--881.

\bibitem[{Vaze et~al.(2022)Vaze, Han, Vedaldi, and
  Zissermann}]{vaze2022openset}
Vaze, S.; Han, K.; Vedaldi, A.; and Zissermann, A. 2022.
\newblock Open-Set Recognition: A Good Closed-Set Classifier is All You Need?
\newblock In \emph{International Conference on Learning Representations
  (ICLR)}.

\bibitem[{Wan et~al.(2024)Wan, Wang, Xie, Li, Huang, and
  Chen}]{wan2024unlocking}
Wan, W.; Wang, X.; Xie, M.-K.; Li, S.-Y.; Huang, S.-J.; and Chen, S. 2024.
\newblock Unlocking the power of open set: A new perspective for open-set noisy
  label learning.
\newblock In \emph{AAAI Conference on Artificial Intelligence}, volume~38.

\bibitem[{Wang et~al.(2022)Wang, Li, Feng, and Zhang}]{wang2022vim}
Wang, H.; Li, Z.; Feng, L.; and Zhang, W. 2022.
\newblock Vim: Out-of-distribution with virtual-logit matching.
\newblock In \emph{Conference on Computer Vision and Pattern Recognition
  (CVPR)}.

\bibitem[{Wang et~al.(2024)Wang, Mu, Zhu, and Hu}]{wang2024exploring}
Wang, Y.; Mu, J.; Zhu, P.; and Hu, Q. 2024.
\newblock Exploring diverse representations for open set recognition.
\newblock In \emph{AAAI Conference on Artificial Intelligence}, volume~38.

\bibitem[{Woo et~al.(2023)Woo, Debnath, Hu, Chen, Liu, Kweon, and
  Xie}]{woo2023convnext}
Woo, S.; Debnath, S.; Hu, R.; Chen, X.; Liu, Z.; Kweon, I.~S.; and Xie, S.
  2023.
\newblock ConvNeXt V2: Co-designing and Scaling ConvNets with Masked
  Autoencoders.
\newblock In \emph{Conference on Computer Vision and Pattern Recognition
  (CVPR)}.

\bibitem[{Xiao et~al.(2010)Xiao, Hays, Ehinger, Oliva, and
  Torralba}]{xiao2010sun}
Xiao, J.; Hays, J.; Ehinger, K.~A.; Oliva, A.; and Torralba, A. 2010.
\newblock Sun database: Large-scale scene recognition from abbey to zoo.
\newblock In \emph{Conference on Computer Vision and Pattern Recognition
  (CVPR)}.

\bibitem[{Xinying~Chen and Hooker(2023)}]{xinying2023guide}
Xinying~Chen, V.; and Hooker, J.~N. 2023.
\newblock A guide to formulating fairness in an optimization model.
\newblock \emph{Annals of Operations Research}, 326(1): 581--619.

\bibitem[{Xu, Shen, and Zhao(2023)}]{xu2023contrastive}
Xu, B.; Shen, F.; and Zhao, J. 2023.
\newblock Contrastive open set recognition.
\newblock In \emph{AAAI Conference on Artificial Intelligence}, volume~37.

\bibitem[{Xu et~al.(2024)Xu, Chen, Franchi, and Yao}]{xu2024scaling}
Xu, K.; Chen, R.; Franchi, G.; and Yao, A. 2024.
\newblock Scaling for Training Time and Post-hoc Out-of-distribution Detection
  Enhancement.
\newblock In \emph{International Conference on Learning Representations
  (ICLR)}.

\bibitem[{Yang et~al.(2020)Yang, Zhang, Yin, Yang, and
  Liu}]{yang2020convolutional}
Yang, H.-M.; Zhang, X.-Y.; Yin, F.; Yang, Q.; and Liu, C.-L. 2020.
\newblock Convolutional prototype network for open set recognition.
\newblock \emph{Transactions on Pattern Analysis and Machine Intelligence
  (TPAMI)}, 44(5).

\bibitem[{Yang et~al.(2022)Yang, Wang, Zou, Zhou, Ding, Peng, Wang, Chen, Li,
  Sun et~al.}]{yang2022openood}
Yang, J.; Wang, P.; Zou, D.; Zhou, Z.; Ding, K.; Peng, W.; Wang, H.; Chen, G.;
  Li, B.; Sun, Y.; et~al. 2022.
\newblock {OpenOOD}: Benchmarking Generalized Out-of-Distribution Detection.
\newblock \emph{Advances in Neural Information Processing Systems (NeurIPS)}.

\bibitem[{Zhang et~al.(2023)Zhang, Yang, Wang, Wang, Lin, Zhang, Sun, Du, Zhou,
  Zhang, Li, Liu, Chen, and Li}]{zhang2023openood}
Zhang, J.; Yang, J.; Wang, P.; Wang, H.; Lin, Y.; Zhang, H.; Sun, Y.; Du, X.;
  Zhou, K.; Zhang, W.; Li, Y.; Liu, Z.; Chen, Y.; and Li, H. 2023.
\newblock Open{OOD} v1.5: Enhanced Benchmark for Out-of-Distribution Detection.
\newblock In \emph{NeurIPS Workshop on Distribution Shifts: New Frontiers with
  Foundation Models}.

\bibitem[{Zhang et~al.(2022)Zhang, Cheng, Zhang, Bonnington, and
  Ge}]{zhang2022learning}
Zhang, X.; Cheng, X.; Zhang, D.; Bonnington, P.; and Ge, Z. 2022.
\newblock Learning Network Architecture for Open-Set Recognition.
\newblock In \emph{AAAI Conference on Artificial Intelligence}, volume~36.

\bibitem[{Zhou et~al.(2017)Zhou, Lapedriza, Khosla, Oliva, and
  Torralba}]{zhou2017places}
Zhou, B.; Lapedriza, A.; Khosla, A.; Oliva, A.; and Torralba, A. 2017.
\newblock Places: A 10 million image database for scene recognition.
\newblock \emph{Transactions on Pattern Analysis and Machine Intelligence
  (TPAMI)}, 40(6).

\bibitem[{Zhou, Ye, and Zhan(2021)}]{zhou_learning_2021}
Zhou, D.-W.; Ye, H.-J.; and Zhan, D.-C. 2021.
\newblock Learning {Placeholders} for {Open}-{Set} {Recognition}.
\newblock In \emph{CVPR 2021}, 4401--4410.

\end{thebibliography}
\clearpage

\end{document}